\documentclass{article}

\usepackage{arxiv}

\usepackage[utf8]{inputenc}
\usepackage[T1]{fontenc}
\usepackage{microtype}
\usepackage{graphicx}
\usepackage{subcaption}
\usepackage{booktabs}
\usepackage{xcolor}
\usepackage{url}
\usepackage{doi}
\usepackage{natbib}
\usepackage{hyperref}
\usepackage{amsmath}
\usepackage{amsfonts}
\usepackage{amssymb}
\usepackage{mathtools}
\usepackage{amsthm}
\usepackage[capitalize,noabbrev]{cleveref}
\usepackage{comment}
\IfFileExists{tcolorbox.sty}{%
  \usepackage[most]{tcolorbox}
  \tcbuselibrary{breakable}
}{%
  \newenvironment{tcolorbox}[1][]{%
    \par\smallskip\noindent\begin{minipage}{\textwidth}\hrule\smallskip
  }{%
    \smallskip\hrule\end{minipage}\par\smallskip
  }
}
\usepackage{xspace}

\newcommand{\rev}[1]{#1}
\newcommand{\orcid}[1]{\href{https://orcid.org/#1}{\includegraphics[scale=0.06]{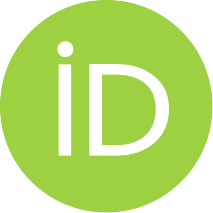}\hspace{1mm}}}

\theoremstyle{plain}

\theoremstyle{definition}

\theoremstyle{remark}

\title{Beyond Inference-Time Search: Reinforcement Learning Synthesizes Reusable Solvers}

\author{%
\orcid{0000-0002-6763-3625}Soheyl Massoudi \\
ETH Z\"urich \\
Z\"urich, Switzerland \\
\texttt{smassoudi@ethz.ch} \\
\And
\orcid{0009-0007-3541-8625}Gabriel Apaza \\
University of Maryland \\
College Park, MD, USA \\
\And
\orcid{0009-0008-1170-2362}Milad Habibi \\
University of Maryland \\
College Park, MD, USA \\
\And
\orcid{0000-0003-3819-8895}Mark Fuge \\
ETH Z\"urich \\
Z\"urich, Switzerland
}

\renewcommand{\shorttitle}{Beyond Inference-Time Search}

\hypersetup{
pdftitle={Beyond Inference-Time Search: Reinforcement Learning Synthesizes Reusable Solvers},
pdfsubject={Machine learning for code, reinforcement learning, combinatorial optimization},
pdfauthor={Soheyl Massoudi, Gabriel Apaza, Milad Habibi, Mark Fuge},
pdfkeywords={large language models, code generation, reinforcement learning, solver synthesis, combinatorial optimization},
}

\begin{document}
\maketitle

\begin{abstract}
\rev{Large language models (LLMs) typically approach combinatorial optimization as an inference-time procedure, solving each instance separately through sampling, search, or repeated prompting. We ask whether reinforcement learning can instead shift part of this reasoning cost into the weights of a code LLM, so that the model synthesizes a reusable solver for an entire problem family. We study this question on Synergistic Dependency Selection (SDS), a controlled variant of constrained Quadratic Knapsack designed to expose a specific failure mode: local signals and strict feasibility constraints make greedy heuristics attractive but unreliable. Under identical scaffolding, Best-of-$64$ base-model sampling saturates at a $\approx 28.7\%$ gap to the global Virtual Best Solver (VBS); code audits show that the base model often retrieves Simulated Annealing templates but misimplements the Metropolis acceptance rule. We fine-tune Qwen2.5-Coder-14B-Instruct with Group Relative Policy Optimization (GRPO) using a feasibility-gated reward and light structural scaffolding. The resulting policy converges to a constraint-aware Simulated Annealing template in 99.8\% of feasible SDS outputs, achieves a 5.0\% gap to that VBS, and is 91$\times$ cheaper in post-generation execution/search cost than cumulative Best-of-$64$ evaluation. A compile-once check shows that one best frozen solver per seed remains highly competitive when reused unchanged across the SDS test set, while an additional-domain evaluation on Job Shop Scheduling provides narrower but positive evidence that the scaffold transfers beyond SDS. Negative ablations reveal the limits of this recipe: standard stabilizers degrade performance, a soft feasibility gate fails, and results remain sensitive to reward normalization and domain-specific design choices.}
\end{abstract}


\section{Introduction}

\rev{The prevailing paradigm in Machine Learning for Code treats the Large Language Model (LLM) as an intelligent agent that solves problems instance-by-instance, either by acting directly as a learned heuristic or by repeatedly translating the problem into executable tool calls \citep{kool_attention_2018, manchanda_gcomb_2020, gao_pal_2023, pan_logic-lm_2023}. In this view, ``reasoning'' is a marginal cost incurred every time a new query is submitted. However, a distinct frontier exists where the model functions not as an interpreter, but as a \textit{compiler}: it analyzes a problem class once and synthesizes a standalone executable that can be reused across the distribution.}

\rev{We investigate the limits of this synthesis capability in the context of recent work on program-space search, amortized solver discovery, and the debate over whether RL mainly sharpens an existing distribution or can also repair behavior that naive sampling does not reliably recover \citep{romera-paredes_mathematical_2024, ye_reevo_2024, novikov_alphaevolve_2025, wang_thetaevolve_2025, yue_does_2025}. Our core question is therefore not only whether RL can improve code generation quality, but whether it can move a code LLM across a qualitatively different regime: from repeated per-instance search to reusable solver synthesis. To test that claim cleanly, we need a domain in which surface template retrieval is insufficient and executable algorithmic logic matters.}

\rev{We use our custom Synergistic Dependency Selection (SDS) benchmark, a strongly NP-hard variant of the Quadratic Knapsack Problem, as that primary stress test. We do not position SDS as a replacement for established Operations Research benchmarks. Instead, we use it as a controlled instrument for deceptive optimization: its rugged reward landscape and strict constraints make it easy to retrieve plausible solver templates yet difficult to operationalize them correctly. It also serves as a cleaner out-of-distribution test of solver synthesis than canonical benchmark families already well represented in model pre-training, reducing the chance that strong behavior can be explained by straightforward benchmark-template recall alone. Under the identical Hero scaffold, the base model still saturates at a 28.7\% gap, so the prompt names a useful reasoning mode but does not operationalize the solver by itself. This lets us ask whether RL can push a code model beyond surface retrieval and into executable search behavior.}

\rev{We focus on three research questions. (1) \textbf{Competence Wall:} Can extensive test-time sampling recover the performance of an RL-trained policy in deceptive landscapes, or does the base model hit a ceiling even when it retrieves superficially relevant templates? (2) \textbf{Semantic Repair:} When RL succeeds, is it merely sharpening the frequency of already-correct solutions, or is it repairing executable algorithmic logic that naive sampling does not reliably recover? (3) \textbf{Amortized Solver Synthesis:} Can training-time optimization distill reasoning cost into a lightweight, reusable program that outperforms repeated per-instance search in the intended deployment regime?}

\rev{To study those questions, we use a feasibility-centric RL protocol that deliberately strips away many common stabilizers (e.g., KL penalties, oracle anchoring, or diversity bonuses). We train a general-purpose coder ({Qwen2.5-Coder-14B}) \citep{hui_qwen25-coder_2024} using GRPO \citep{shao_deepseekmath_2024} with a simple feasibility-gated reward and light structural scaffolding. This deliberately minimalist setup is useful precisely because it lets us isolate what such a recipe is already sufficient to produce.}

\paragraph{Contributions.}
\rev{Our contributions are empirical and mechanistic rather than algorithmic. First, we identify a \textbf{competence wall}: under identical prompting, Best-of-$64$ base-model sampling saturates at a $28.7\%$ SDS gap to VBS even though the model often retrieves superficially appropriate solver templates. Second, we show a \textbf{semantic-repair effect}: RL drives our learned policy, \textbf{Hero}, toward an almost ubiquitous constraint-aware Simulated Annealing template ($99.8\%$ convergence on feasible SDS outputs) and repairs the executable logic that base-model samples frequently misimplement. This is not explained away by a larger one-shot search over base samples alone: even an adaptive tournament across 191,699 unique base codes still fails to recover a reusable solver close to Hero on SDS. Third, we directly validate \textbf{amortized solver synthesis}: the learned policy occupies a distinct Pareto region on SDS ($5.0\%$ gap to VBS, $6.3\times$ faster than CP-SAT in post-generation execution cost), and a frozen compile-once evaluation shows that one generated solver per seed remains highly competitive when reused unchanged across the SDS test set. The same prompt alone does not produce this behavior: Base Best-of-$64$ still plateaus badly under the Hero scaffold, whereas RL makes the named procedure executable on SDS. Additional JSSP experiments provide narrower evidence that the scaffold extends beyond SDS, while negative ablations make clear that the current recipe still depends on explicit feasibility-first design choices.}

\section{Related Work}
\paragraph{From Neural Solvers to Solver Synthesis.}
\rev{Traditional Neural Combinatorial Optimization (NCO) trains models to act as black-box heuristics, learning a direct mapping from instances to solutions ($x = f_\theta(c)$), \citep{kool_attention_2018, manchanda_gcomb_2020}. These methods function as \textit{interpreters}, requiring the neural network to be active during inference for every new problem instance. In contrast, we frame optimization as a code synthesis task ($p = \pi_\theta(c)$). Our policy acts as a \textit{compiler}: it generates an explicit algorithmic artifact $p$ that can be inspected, executed, and potentially reused without keeping the neural network in the inference loop. Our novelty claim is therefore not a new NCO architecture, but a direct empirical study of when a code LLM can cross from per-instance solving into reusable solver synthesis.}

\paragraph{Reinforcement Learning for Code.}
\rev{Using execution feedback to ground LLMs is an emerging frontier. While approaches like CodeRL \citep{le_coderl_2022}, RLTF \citep{liu_rltf_2023}, RLEF \citep{gehring_rlef_2025}, and AutoTriton \citep{li_autotriton_2025} optimize for binary correctness or kernel latency, we target the \textit{mathematical quality} of the produced solver inside a deceptive optimization landscape. We do not claim a new RL algorithm. Rather, we use a deliberately simple GRPO-based recipe to study whether sparse feasibility-first feedback can change the executable logic of generated solvers in a way that naive sampling does not.}

\paragraph{Neuro-Symbolic Reasoning.}
\rev{Frameworks like PAL \citep{gao_pal_2023} and Logic-LM \citep{pan_logic-lm_2023} delegate computation to an external Python interpreter or symbolic solver. These methods typically treat the LLM as a translator and the solver as a fixed black box. Our approach diverges by training the LLM to generate the solver itself, including the algorithmic logic for constraint verification and search. This makes the produced artifact a white-box object that can be audited for both failure and repair, which is central to our competence-wall and semantic-repair analysis.}

\paragraph{Generative Hyper-Heuristics.}
\rev{Modern hyper-heuristics use LLMs to search the program space, often employing evolutionary strategies in the prompt space (e.g., FunSearch \citep{romera-paredes_mathematical_2024}, ReEvo \citep{ye_reevo_2024}, AlphaEvolve \citep{novikov_alphaevolve_2025}) or test-time training loops (e.g., ThetaEvolve \citep{wang_thetaevolve_2025}). While powerful, these methods generally rely on extensive inference-time search, incurring high marginal costs per instance. Our comparison point is therefore not simply ``better numbers,'' but a different compute allocation: reasoning effort is pushed offline into policy weights so that the deployment object is a reusable solver rather than a fresh search process for every instance.}

\paragraph{The Reasoning Capacity Debate.}
The nature of LLM ``reasoning'' remains contested, often framed as a tension between approximate retrieval and genuine planning \citep{huang_large_2024, valmeekam_planbench_2023}. Recently, \citet{yue_does_2025} argued that RL with Verifiable Rewards (RLVR) primarily acts as a ``distribution sharpener'', effectively just improving the efficiency of finding solutions that the base model could already access via sampling. Our work directly engages with this hypothesis by investigating \textit{deceptive optimization} landscapes\textemdash environments where immediate gradient signals mislead heuristics away from the global optimum. This serves as a critical test case for whether RL merely filters existing solutions or synthesizes algorithmic behavior that \rev{the base model does not reliably reach via naive sampling on this benchmark.}

\section{Problem Formulation}

\rev{We focus on Synergistic Dependency Selection (SDS), a constrained variant of the Quadratic Knapsack Problem (QKP). Unlike the standard Knapsack Problem which is solvable in pseudo-polynomial time, SDS generalizes the Maximum Independent Set problem \citep{karp_reducibility_2010}, rendering it strongly NP-hard \citep{pisinger_quadratic_2007}. We use SDS as a controlled stress test for solver synthesis under deceptive local signals, for two strategic reasons:}

\begin{enumerate}
    \item \rev{\textbf{Out-Of-Distribution (OOD) Challenge:} Unlike canonical problems (e.g., TSP, Standard Knapsack) present in pre-training, SDS presents a novel task structure. This reduces the chance that strong performance can be explained by straightforward template recall alone, and instead pressures the model to compose search and constraint-handling primitives into executable logic.}
    \item \rev{\textbf{Deceptive Landscapes:} We generate instances where quadratic interaction terms dominate linear weights ($|W_{ij}| \gg |w_i|$). This creates a landscape where local gradients actively contradict global optimality, making SDS a useful test of whether RL can repair solver behavior beyond the greedy or superficially plausible trajectories favored by next-token prediction.}
\end{enumerate}

\subsection{Formal Definition}
Let $x \in \{0, 1\}^n$ be the binary selection vector. The goal is to maximize total utility $S(x)$, defined as the sum of linear weights $w_i$ (intrinsic value) and pairwise quadratic interactions $W_{ij}$ (synergistic/antagonistic effects):
\begin{equation}
\text{Maximize } S(x) = \sum_{i=0}^{n-1} w_i x_i + \sum_{(i,j) \in \mathcal{E}} W_{ij} x_i x_j.
\end{equation}
A solution is valid if and only if it satisfies the strict constraints $\mathcal{C}$:
(1) \textbf{Cardinality:} Selection size is bounded by $L \le \sum_{i=0}^{n-1} x_i \le U$;
(2) \textbf{Precedence:} Dependent items require prerequisites ($\forall (i, j) \in \mathcal{P}, x_j \le x_i$);
(3) \textbf{Mutual Exclusion:} Conflicting pairs are exclusive ($\forall (a, b) \in \mathcal{M}, x_a + x_b \le 1$); and
(4) \textbf{Group Limits:} At most one item per group $g \in \mathcal{G}$ ($\sum_{i \in g} x_i \le 1$).

\section{Methodology}
\label{sec:methodology}

\rev{We present \textbf{Hero}, a policy fine-tuned via Group Relative Policy Optimization (GRPO) \citep{shao_deepseekmath_2024}. Hero is intentionally minimalist: it drops oracle anchoring and diversity bonuses in favor of \textit{scaffolded verification}\textemdash a hierarchical reward structure that separates executable structure from objective quality. Hero also uses a scaffolded system prompt whose core loop is Deconstruct-Hypothesize-Critique; see Appendix~\ref{app:prompts}. We emphasize this simplicity because the scientific question is not whether a heavily engineered recipe can work, but what a comparatively lean feasibility-first recipe is already sufficient to induce.}

\subsection{Policy Optimization}
Following \citet{yu_dapo_2025}, we use a group size $G=64$ with no KL regularization ($\beta=0.0$). GRPO stabilizes updates by standardizing rewards within each sampled group, enabling the model to discern relative policy improvements even when absolute rewards remain sparse.

\subsection{Hierarchical Reward Design}
We define a composite reward $r(o)$ to enforce functional priority:
\begin{equation}
    r(o) = \lambda_1 R_{\text{format}} + \lambda_2 R_{\text{exec}} + \lambda_3 R_{\text{nom}}
\end{equation}
where $\lambda = (0.1, 0.2, 0.7)$. This weighting ensures structural compliance bootstraps the policy without overriding the primary optimization objective.

\subsubsection{Structural Scaffolding}
The auxiliary terms $R_{\text{format}}$ and $R_{\text{exec}}$ function as semantic type-checks. $R_{\text{format}}$ ensures the model follows a reasoning-action trace ($\langle\text{think}\rangle \dots \langle\text{code}\rangle$). $R_{\text{exec}}$ provides dense feedback on code validity:
\begin{equation}
    R_{\text{exec}} = R_{\text{syntax}} + R_{\text{schema}} + R_{\text{struct}} + R_{\text{feas}}
\end{equation}
A pivotal element is an explicit anti-lazy sorting penalty applied within $R_{\text{exec}}$. This penalizes the pre-trained prior's tendency to adopt linear greedy heuristics (e.g., sorting by weight) while ignoring quadratic interaction terms. Separately, the positive graph-structure bonus inside $R_{\text{struct}}$ is subject to curriculum fading $\alpha(t)$ as training progresses.

\subsubsection{Feasibility-Gated Optimality}
The nominal reward, $R_{\text{nom}}$, is the primary signal for objective maximization but is subject to a feasibility cliff:
\begin{equation}
    R_{\text{nom}} = 
    \begin{cases} 
    \text{Norm}(\text{Score}(o)) & \text{if } N_{\text{vio}} = 0 \\
    0 & \text{if } N_{\text{vio}} > 0
    \end{cases}
\end{equation}
where $N_{\text{vio}}$ is the count of constraint violations. This gating forces the policy to prioritize hard constraint satisfaction (e.g., mutex or precedence) before optimizing the objective value. Details in Appendix~\ref{app:rewards}.

\section{Experiments}
\label{sec:experiments}

\subsection{Setup}

\paragraph{Models and Training.}
We initialize our policy with Qwen2.5-Coder-14B-Instruct \citep{hui_qwen25-coder_2024}, selected for its robust base coding and reasoning performance despite lacking native test-time compute scaling. Preliminary experiments with the 7B variant yielded insufficient policy stability, necessitating the 14B capacity for complex SDS tasks. To evaluate sample efficiency, we train Hero for a restricted budget of 90 global steps ($\approx$4 hours) on 12 NVIDIA GH200 GPUs. Due to memory constraints (4 prompts per step), the model observes only 360 prompts, or about 4.5\% of a full prompt-level epoch ($\approx 8{,}000$ prompts), while still synthesizing 23,040 reasoning traces. This regime specifically evaluates the capacity for \textit{deep exploration} over broad exposure.

\paragraph{The Syndeopt Benchmark.}
\rev{We evaluate on \textbf{Syndeopt}, a controlled SDS stress-test benchmark of 10,000 instances stratified to penalize greedy behavior. We use it as a scientific instrument rather than as a replacement for standard OR suites: the goal is to isolate settings where plausible template retrieval is easy but executable search logic is hard to operationalize correctly. It includes \textit{Dense Deceptive} instances, where strong interactions ($|W_{ij}| \gg |w_i|$) trap gradient-following heuristics, and \textit{Structural Traps} involving deep dependency chains. We use a held-out set of 1,000 instances for evaluation.}

\paragraph{Baselines.}
\rev{Hero is compared against four baseline families that answer different scientific questions. \textbf{Naive-sampling baselines} test the competence-wall hypothesis using the base model under Best-of-$N$. \textbf{Frontier test-time-search baselines} such as ShinkaEvolve~\citep{lange_shinkaevolve_2025} represent stronger program-space search at deployment time. \textbf{Reusable-solver baselines} such as our manually specified Simulated Annealing implementation test whether Hero only recovers what a competent human would already write. Finally, \textbf{classical optimization baselines} including Greedy Search, Local Search, a time-limited Branch-and-Bound heuristic, and CP-SAT~\citep{perron_cp-sat-lp_2023} anchor solution quality and runtime against standard non-neural methods. All methods are subject to a 5-second execution limit per instance.}

\paragraph{Metrics.} 
Success is measured via three primary metrics:
\begin{enumerate}
    \item \textbf{Pass Rate:} The percentage of instances yielding a strictly feasible solution ($N_{\text{vio}}=0$).
    \item \textbf{Gap to VBS:} $(\text{VBS} - \max(0, S)) / \text{VBS}$, where $S$ is the achieved score (clipped to non-negative) and $\text{VBS}$ is the Virtual Best Solver score. Here VBS is the best feasible score observed across all compared methods, not a certified distance to the true optimum. To decouple validity from optimization quality, mean gaps in Table~\ref{tab:main_results} are computed \textit{conditionally} on feasibility. For global robustness profiles (Fig.~\ref{fig:robustness}), infeasible solutions are assigned a gap of 1.0.
    \item \textbf{Computational Cost:} Total CPU-seconds excluding LLM inference. For the Base model, this is the cumulative per-instance search-and-evaluation cost across sampled candidates ($\sum_{i=1}^{N} t_i$ for Best-of-$N$). For Hero and ShinkaEvolve, it is single-solver execution cost after code generation ($t_1$). For CP-SAT, it reflects aggregate parallel load.
\end{enumerate}

Comprehensive details regarding the experimental protocol and implementation are provided in Appendices~\ref{app:experimental_details} and \ref{app:implementation}.

\subsection{Main Results}

\begin{figure}[t!]
\centering
    \begin{subfigure}[t]{0.48\textwidth}
        \centering
        \includegraphics[width=\textwidth]{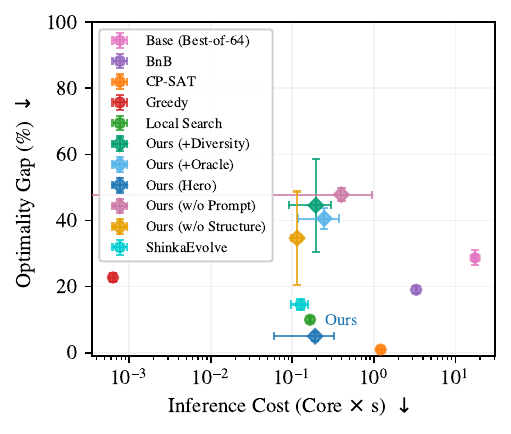}
        \caption{\textbf{Efficiency Frontier.} Hero dominates the Pareto frontier \textemdash second only to CP-SAT, achieving lower gaps to VBS than the refreshed ShinkaEvolve deployment-time search baseline. Note the log-scale x-axis: our specialized policy is $\approx 91 \times$ cheaper than the Base Model \rev{when comparing solver execution against cumulative Best-of-64 sampling cost}. Gap to VBS is reported only on feasible solutions.}
        \label{fig:efficiency}
    \end{subfigure}
    \hfill
    \begin{subfigure}[t]{0.48\textwidth}
        \centering
        \includegraphics[width=\textwidth]{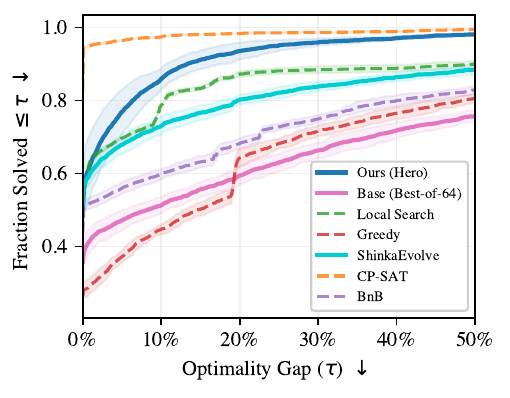}
        \caption{\textbf{Robustness Profile.} Cumulative fraction of problems with gap to VBS $\leq \tau$. Our method (Blue) dominates the upper-left region, indicating high reliability. Infeasible solutions are included with gap $=100\%$.}
        \label{fig:robustness}
    \end{subfigure}
    \caption{\textbf{Main Performance Analysis.} Aggregated results across 3 seeds ($N=3000$ instances). Error bars and shaded regions indicate standard deviation.}
    \label{fig:main_performance}
\end{figure}

\rev{We evaluate our method across three independent seeds (101, 202, 303). Figure~\ref{fig:main_performance} summarizes the aggregate evidence for the three main claims: a competence wall for naive sampling, semantic repair under RL, and amortized solver synthesis through reusable generated code.}

\paragraph{The Efficiency Frontier.} \rev{As shown in Figure~\ref{fig:efficiency}, Hero occupies a distinct Pareto-optimal region. It achieves a $5.0 \pm 1.3$\% gap to VBS, lower than Local Search ($10.0 \pm 0.7$\%), while also outperforming the refreshed ShinkaEvolve baseline that uses strong proprietary mutation/search models at deployment time~\citep{openai_gpt5_2025}. We interpret this not simply as ``RL wins on SDS,'' but as evidence that a specialized policy can amortize search into a reusable solver artifact more effectively than repeated deployment-time search on this benchmark. For generated-solver methods such as Hero and ShinkaEvolve, the plotted times measure solver execution after code generation. For Base Best-of-64, the plotted cost is instead the cumulative per-instance execution/selection cost across the sampled candidate programs, still excluding LLM inference. We therefore interpret these timings as amortized execution/search costs rather than end-to-end deployment latency. A direct compile-once check strengthens that interpretation: the frozen Hero solver remains highly competitive when reused unchanged across the test set, and the freshest timing comparison shows that reusable compiled solvers are far cheaper than repeated Best-of-64 search in this amortized setting.}

\paragraph{Stochastic Dominance.} \rev{The performance profile (Figure~\ref{fig:robustness}) shows that the generated Hero solver family dominates the greedy-style baselines over a wide range of tolerances. The sharp rise near $0\%$ gap to VBS indicates that RL is not merely increasing feasibility; it is concentrating mass on near-optimal executable behavior.}

\begin{table}[h]
\centering
\caption{\textbf{Aggregated Results (Mean $\pm$ Std).} Comparison across 3 seeds. Our Hero configuration achieves 97.8\% pass rate and a 5.0\% gap to VBS, surpassing standard heuristics and approaching the strongest time-limited classical reference (CP-SAT). Gap to VBS is calculated only on feasible solutions using the global VBS (maximum feasible score across all compared methods). \textsuperscript{*}Time denotes Computational Cost (CPU-sec): aggregate parallel load for CP-SAT, cumulative sampled-candidate search/evaluation cost for Base selection, and single-solver execution for Hero/ShinkaEvolve. \rev{LLM inference is excluded from this column for all methods, so these values should be interpreted as post-generation amortized execution/search costs rather than end-to-end wall-clock costs.}}
\label{tab:main_results}
\small
\setlength{\tabcolsep}{8pt}
\begin{tabular}{lccc}
\toprule
Method & Pass Rate (\%) $\uparrow$ & Gap to VBS (\%) $\downarrow$ & Time (s) $\downarrow$ \textsuperscript{*} \\
\midrule
Ours (Hero) & $97.8_{\pm 0.2}$ & $5.0_{\pm 1.3}$ & $0.191_{\pm 0.131}$ \\
Base (Best-of-64) & $85.6_{\pm 1.3}$ & $28.7_{\pm 2.3}$ & $17.434_{\pm 0.793}$ \\
Ours (+Oracle) & $65.3_{\pm 11.5}$ & $40.6_{\pm 3.1}$ & $0.247_{\pm 0.128}$ \\
Ours (+Diversity) & $61.3_{\pm 9.8}$ & $44.6_{\pm 14.0}$ & $0.195_{\pm 0.103}$ \\
Ours (w/o Structure) & $74.8_{\pm 19.2}$ & $34.7_{\pm 14.1}$ & $0.115_{\pm 0.018}$ \\
Ours (w/o Prompt) & $64.4_{\pm 7.8}$ & $47.8_{\pm 1.9}$ & $0.400_{\pm 0.548}$ \\
ShinkaEvolve & $91.6_{\pm 1.2}$ & $14.4_{\pm 2.0}$ & $0.127_{\pm 0.029}$ \\
CP-SAT & $\mathbf{98.3_{\pm 0.3}}$ & $\mathbf{1.0_{\pm 0.0}}$ & $1.212_{\pm 0.054}$ \\
Local Search & $\mathbf{98.3_{\pm 0.3}}$ & $10.0_{\pm 0.7}$ & $0.165_{\pm 0.005}$ \\
Greedy & $\mathbf{98.3_{\pm 0.3}}$ & $22.8_{\pm 1.3}$ & $\mathbf{0.001}$ \\
BnB & $97.3_{\pm 0.4}$ & $19.1_{\pm 1.1}$ & $3.301_{\pm 0.048}$ \\
\bottomrule
\end{tabular}
\end{table}

\begin{table}[h]
\centering
\caption{\textbf{Compile-Once Evidence for Reusable Solvers.} Standard Hero regenerates one solver per instance, whereas Frozen Hero and Manual SA execute one fixed solver unchanged across the full SDS test split. Base Best-of-64 re-runs cumulative multi-sample search per instance. Pass rate and gap use the fixed-code comparison bundle; wall-clock totals are representative seed101 evaluation totals.}
\label{tab:compile_once_main}
\small
\begin{tabular}{lccc}
\toprule
Method & Pass (\%) & Gap (\%) & Wall-Clock (s) \\
\midrule
Standard Hero & 97.83 & 4.82 & 2731 \\
Frozen Hero & 97.87 & 4.34 & 2342 \\
Manual SA & 95.63 & 5.32 & 3103 \\
Base Best-of-64 & 85.6 & 28.7 & 25480 \\
\bottomrule
\end{tabular}
\end{table}

\paragraph{Robustness Across Seeds.}
\rev{Our method is also consistent across all three seeds (Table~\ref{tab:main_results}). With a standard deviation of only 0.2\% for pass rate and 1.3\% for gap to VBS, the result is not driven by a single lucky run. This matters for interpretation: the competence-wall and semantic-repair story is visible at the level of repeated training outcomes, not only within one selected checkpoint.}

\paragraph{Efficiency vs. Optimality.}
\rev{While CP-SAT achieves a near-zero gap to VBS, Hero trades only a modest 5.0\% gap to VBS for a 6.3$\times$ improvement in post-generation execution cost relative to CP-SAT and dramatically lower sampled-candidate evaluation cost than Base Best-of-64. A strict compile-once check confirms solver reuse: one frozen Hero solver per seed reaches 97.87\% pass rate and 4.34\% mean gap when executed unchanged across the SDS test set (Appendix~\ref{app:baseline_eval_additions}, Table~\ref{tab:compile_once_main}).}

\subsection{Ablation: The Necessity of Structural Scaffolding}

\rev{To isolate what is actually load-bearing in the recipe, we compare Hero against both ``High-Interference'' (Oracle/Diversity) and ``Zero-Scaffold'' (Pure Minimalist) baselines. The goal of this ablation suite is not only to show that alternatives perform worse, but to identify which ingredients drive competence-wall crossing and semantic repair. Results in the bottom section of Table~\ref{tab:main_results} reveal that effective learning requires a delicate balance: strong feasibility-first guardrails without dense outcome shaping. We reserve that table for the primary \emph{configuration ablations} that directly change the training recipe. Additional matched studies are still important, but they play different roles: the soft gate is a matched alternative reward configuration, the normalization study is a matched sensitivity analysis on one load-bearing Hero component, and the feasibility-sparsity and timeout analyses are diagnostics of why the recipe behaves as it does.}

\paragraph{Hero Configuration (Scaffolded Verification).}
The Hero configuration uses a hierarchical reward designed to distinguish valid syntax from valid reasoning:
\begin{itemize}
    \item {Format \& Syntax ($30\%$):} Strict formatting and execution checks to ensure valid Python generation.
    \item {Structural Scaffolding ($R_{\text{exec}}$):} A critical guardrail that detects graph-theoretic awareness. This component applies a negative penalty to ``lazy sort'' heuristics (sorting by weight while ignoring interactions), explicitly suppressing the base model's greedy prior.
    \item {Nominal Optimality ($70\%$):} The feasibility-gated objective value, devoid of relative oracle shaping.
\end{itemize}

\paragraph{Why Complexity Failed.}
Adding standard RL stabilizers consistently degraded performance.
\textbf{Oracle Anchoring ($65.3\%$ pass):} While intended to guide optimization, the relative reward created a fatal \textit{signal conflict}. The high variance of the oracle signal drowned out the strict feasibility gate, teaching the model to gamble on high-scoring but invalid solutions.
\textbf{Diversity Penalties ($61.3\%$ pass):} Penalizing low entropy failed because valid algorithmic templates are \textit{sparse} in highly constrained domains. Forcing diversity actively pushed the policy away from the valid Simulated Annealing template into infeasible syntax regions.

\paragraph{Why Pure Minimalism Failed.}
Most critically, the \textbf{w/o Structure} ablation reveals that removing scaffolding leads to an ``Implementation Wall''. While the policy attempted meta-heuristics (Simulated Annealing) in two seeds out of three, it failed to adapt them to the constrained landscape. 
Code analysis reveals a brittle pattern: passive constraint filtering (see comparative code analysis in Appendix~\ref{app:implementation_wall}). Unlike the Hero policy, which more often synthesized robust constraint verification (using either active retry loops or tuned rejection sampling), the minimalist policy simply discarded invalid neighbors while continuing to cool the temperature. This resulted in \textit{brittle} searches that, lacking the hyperparameter alignment of the Hero policy, exhausted their compute budget on invalid states.
Combined with a $34.5\%$ rate of invalid greedy solutions (the pre-training prior), this confirms that structural scaffolding is essential not just for algorithm selection, but for the \textit{robust operationalization} of those algorithms.

\paragraph{Guidance and Failure Modes.}
The system prompt acts as the second pillar of our scaffold. Removing it (``w/o Prompt'') caused a unique spike in runtime errors to $29.8\%$ (Table~\ref{tab:error_types}). We therefore interpret this as an ablation of the full deployed prompt package, not a perfectly isolated cognitive-scaffolding intervention: the minimal prompt also removes some environment reminders, so the result is consistent with prompt-level scaffolding being important for generating executable Python.
The failure mode analysis (Figure~\ref{fig:failure_modes} in Appendix~\ref{app:results}) reveals distinct error profiles: the Hero configuration fails exclusively via timeouts ($2.2\%$) with zero logic errors, indicating structurally sound but computationally expensive code. In contrast, the ``w/o Structure'' ablation suffers primarily from constraint violations ($34.5\%$). This confirms that our structural scaffold does not merely slow the model down—it fundamentally enforces the constraint-satisfaction barrier that the minimalist ablation fails to respect.
\rev{We also evaluated a soft-gated nominal reward that replaces Hero's hard feasibility cliff with a graded penalty on simulator-reported violations while keeping the rest of the recipe fixed. Concretely, the normalized nominal score is reduced by a fixed per-violation penalty and clipped to $[0,1]$, instead of collapsing infeasible samples directly to zero. Across seeds 101/202/303, this soft-gate variant underperforms canonical Hero on both feasibility (57.4\% vs.\ 97.8\% average pass rate) and shared-VBS mean gap (43.50\% vs.\ 4.09\%). Because infeasible solutions incur 100\% gap under this shared-VBS comparison, the gap degradation is partly driven by the sharp increase in infeasible outputs, which is why we interpret pass rate and gap jointly. Its dominant failure mode remains precedence violations. This negative result suggests that, on SDS, the hard gate is not merely an overly sparse signal; it provides a useful feasibility-first training pressure.}
\rev{We treat the remaining matched analyses differently on purpose. The normalization experiment is not a separate row in Table~\ref{tab:main_results} because it is not a new recipe family like ``+Oracle'' or ``w/o Structure''; it is a matched sensitivity study that perturbs one specific Hero design choice while keeping the rest of the stack fixed. Likewise, the feasibility-sparsity and timeout studies are diagnostics rather than ablations: they explain whether the hard gate starves GRPO of signal and where Hero's residual failures concentrate. Appendix~\ref{app:ablation_rewards} organizes these three experiment types explicitly so that the paper reads as one coherent evaluation package rather than a sequence of add-ons.}

\begin{figure}[h]
\centering
\includegraphics[width=0.95\textwidth]{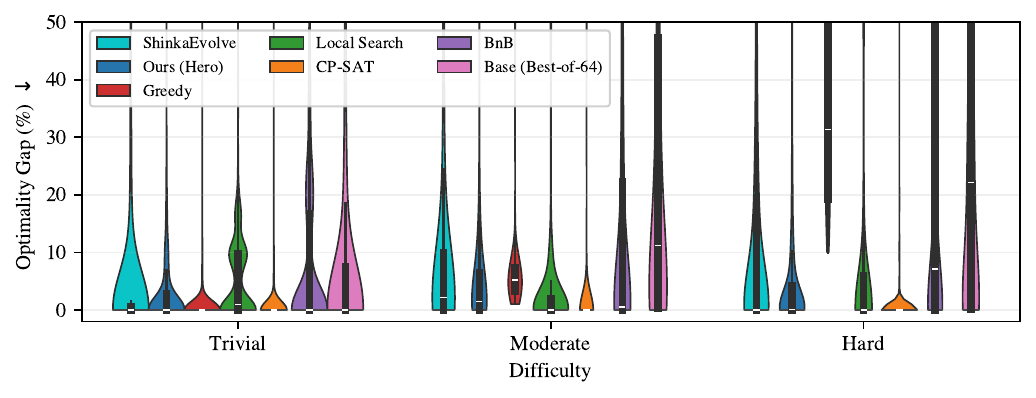}
\caption{\textbf{Performance Scaling by Difficulty.} Distribution of gaps to VBS across Trivial, Moderate, and Hard instances. While baselines degrade on Hard problems (red/green boxes shift up), our method (blue) maintains low gaps to VBS, validating the emergence of robust search heuristics. Infeasible solutions are included with gap $=1.0$.}
\label{fig:stratified}
\end{figure}

\paragraph{Difficulty Scaling.}
Figure~\ref{fig:stratified} decomposes performance by problem difficulty. Instances are classified as \textit{Trivial}, \textit{Moderate}, or \textit{Hard} based on the relative performance gap between a greedy baseline and the Virtual Best Solver (VBS): Trivial instances have $<1\%$ gap (greedy is nearly optimal), Moderate have $1$--$10\%$ gap, and Hard have $\geq 10\%$ gap (greedy fails significantly). On \textit{Trivial} instances, all methods perform well. However, on \textit{Hard} instances (characterized by deceptive landscapes), greedy heuristics collapse (median gap $>30\%$). Figure~\ref{fig:stratified} shows the Base Model's performance variance explodes on Hard instances, whereas the RL policy remains tight. In contrast, our RL-trained policy maintains a median gap $<5\%$, demonstrating that the learned search strategies generalize to unseen structural complexity.

\subsection{Robustness and Generalization}
\label{sec:robustness}

\paragraph{No Catastrophic Forgetting.}
A common failure mode in RL fine-tuning on narrow tasks is overfitting to the target syntax at the expense of general capabilities. To quantify this risk, we evaluated our checkpoints on \textbf{HumanEval} \citep{chen_evaluating_2021} and \textbf{MBPP} \citep{austin_program_2021} using the BigCode Evaluation Harness. We utilized greedy decoding ($T=0, N=1$) to ensure strict comparability with the Qwen2.5-Coder technical report \citep{hui_qwen25-coder_2024}.

As shown in Appendix~\ref{app:results} (Table~\ref{tab:bigcode_results}), our Hero policy achieves an $80.9\%$ Pass@1 on HumanEval, statistically tied with the Base model ($82.3\%$) given the variance ($\pm 0.8$). This confirms that our scaffolded training paradigm acts as a \textit{capability sharpener}: the model learns to context-switch effectively, generating strict SDS solvers when prompted with the SDS scaffold, while retaining standard Python fluency for general tasks.

\paragraph{Evaluation Beyond SDS.}
\rev{To test whether the solver-synthesis recipe extends beyond SDS, we applied the same GRPO setup, reward-stack structure (format / execution / nominal with weights 0.10 / 0.20 / 0.70), and solver-writing prompt philosophy to JSSP, a standard Job Shop Scheduling Problem benchmark. Under one frozen JSSP training/evaluation regime and the same final-checkpoint rule for all three seeds, Hero attains $99.3\%\pm1.2\%$ pass rate and $6.12\%\pm3.80\%$ mean gap to a JSSP VBS over 3{,}000 held-out instances. Across all three seeds, Hero outperforms the heuristic JSSP baselines (Local Search, Most Work Remaining (MWKR), Most Operations Remaining (MOPR), Shortest Processing Time (SPT), and Longest Processing Time (LPT)) and trails only OR-Tools. We treat JSSP as additional evidence of cross-domain viability for the scaffold, not as proof of domain-agnostic transfer: the high-level training recipe carries over, while the simulator, solver contract, and light semantic shaping remain domain-specific. Appendix~\ref{app:transfer_domains} details this setup.}

\begin{figure}[h]
\centering
\includegraphics[width=0.55\textwidth]{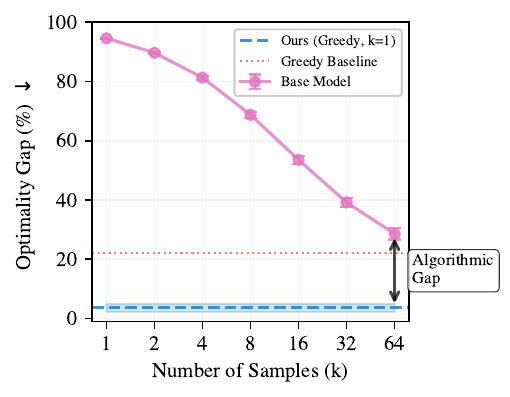}
\caption{\textbf{Scaling Limitations of the Base Model.} Gap to VBS vs.\ sample count $k$ ($\log_2$ scale). Even with the full Hero prompt, the Base model saturates at $\approx 28.7\%$ gap to VBS, showing severe diminishing returns. The plotted curve uses the unconditional SDS gap metric, so subsamples with no feasible solution contribute $100\%$ gap. In contrast, our RL policy (dashed line, $k=1$) achieves $5.0\%$ gap via greedy decoding, far surpassing the constructive Greedy baseline (dotted). This ``Algorithmic Gap'' \rev{indicates that, on SDS, RL reaches solver behavior that naive sampling does not recover in practice}.}
\label{fig:scaling_gap}
\end{figure}

\paragraph{Latent Capacity Limit.} \rev{We next test whether RL merely distills latent knowledge. Recent work argues that RLVR primarily acts as a ``distribution sharpener,'' improving sampling efficiency without expanding the reasoning boundary \citep{yue_does_2025}. To probe that claim, we evaluate the Base model using Best-of-$N$ sampling ($N=64$) with the identical scaffolded ``Hero'' prompt. As shown in Figure~\ref{fig:scaling_gap}, the Base model saturates at a $\approx 28.7\%$ gap to VBS. To diagnose this competence wall, we audited the same raw Base Best-of-$64$ code pool used by universal search: 192,000 generations collapse to 191,699 unique extracted codes across the three seeds (Appendix~\ref{app:base_code_audit}). The Base model \emph{does} access the right surface family at a superficial level: 21.9\% of those unique codes are SA-like under a structural heuristic. However, presence $\neq$ competence. Within that SA-like subset, 28.8\% exhibit the \textit{global-best} acceptance bug, and among the remaining SA-like codes only 23.8\% appear structurally complete under a coarse static audit. This is also not an artifact of inspecting only a small slice of the search pool. To search more exhaustively for a correct reusable solver, we conducted an adaptive tournament across the same 191,699 unique base codes (Appendix~\ref{app:universal_search}). The single best code still achieved only a 24.04\% gap to VBS ($5\times$ worse than Hero's 5.0\%). On SDS, the combined audit and tournament results strengthen the case that RL is doing more than distribution sharpening: it appears to repair the executable logic required for valid optimization.}

\subsection{Case Study: Semantic Repair in Generated Solvers}
\label{sec:emergent_discovery}

\begin{figure}[t!]
\centering
\begin{tcolorbox}[colback=red!5!white, colframe=red!50!black, title=\textbf{Base Model Failure (Logic Hallucination)}, fonttitle=\bfseries, arc=0mm]
\scriptsize
\begin{verbatim}
# ... [Standard setup matches Hero] ...
    # LOGIC HALLUCINATION: Global Best Bug
    # Compares neighbor against BEST_value,
    # not CURRENT_value.
    if neighbor_value > best_value or \
       random.random() < math.exp( \
           (neighbor_value - best_value) / T):
        
        solution = neighbor_solution[:] 
        if neighbor_value > best_value:
            best_value = neighbor_value 
            
    # Result: The search "freezes" once a
    # local optimum is found.
\end{verbatim}
\end{tcolorbox}
\vspace{-2mm} 
\begin{tcolorbox}[colback=blue!5!white, colframe=blue!50!black, title=\textbf{Hero Success (Correct Physics)}, fonttitle=\bfseries, arc=0mm, top=0mm]
\scriptsize
\begin{verbatim}
# ... [Initialization of T=1000, cooling=0.99] ...
    # Constraint Guard (Active Verification)
    while not is_feasible(neighbor):
        idx = random.randint(0, n_vars - 1)
        neighbor[idx] = not neighbor[idx]
    
    # CORRECT PHYSICS: Metropolis Criterion
    # Compares neighbor against CURRENT (Delta).
    n_score = calculate_score(neighbor)
    delta = n_score - current_score
    
    if delta > 0 or \
       random.random() < math.exp(delta/T):
           current_sol = neighbor 
           current_score = n_score
    
    # Result: The search correctly anneals
    # to escape local traps.
\end{verbatim}
\end{tcolorbox}
\caption{\textbf{Visualizing the Logic Hallucination Gap.} Both models retrieve the Simulated Annealing template. However, the Base Model (Top) hallucinates the acceptance logic, comparing neighbors to the running global best found so far rather than the current state, which effectively freezes exploration once the search reaches a local maximum. The Hero policy (Bottom) synthesizes the correct physics ($\Delta = S_{\text{new}} - S_{\text{curr}}$) and active constraint guards, enabling functional optimization.}
\label{fig:logic_gap}
\end{figure}

\rev{To investigate whether the model merely memorizes surface-level patterns or actually assembles functional optimization algorithms, we analyzed the Hero solutions across the full test set ($N=1000$ per seed). The main question in this section is not philosophical ``discovery,'' but whether RL changes the executable behavior of retrieved solver templates in a way that explains the competence-wall gap.}

\paragraph{Algorithmic Convergence.}
\rev{We quantified structural convergence using static analysis on $N=2,935$ feasible solutions aggregated across three seeds. We found that 99.8\% $\pm$ 0.3\% of generated solutions adhere to a functionally equivalent template: Simulated Annealing with a ``Constraint Guard'' mechanism (using either active \texttt{while} loops or strict conditional filtering). This near-perfect convergence indicates that the RL process does not merely increase solution diversity or scatter across many heuristics; it concentrates on a reusable solver family that is both simple enough to emit reliably and expressive enough to survive deceptive landscapes. We hypothesize that this convergence is helped by an inductive bias toward compact search programs: compared with Genetic Algorithms or Tabu Search, Simulated Annealing offers high optimization leverage per token of generated code and per millisecond of execution time.}

\paragraph{The Logic Hallucination Gap.} \rev{Why does the Base Model's SA fail (28.7\% gap) while Hero's succeeds (5.0\% gap) despite using the same high-level template? We compared the Base-model SA-like code pool against Hero and identified a substantial \textbf{Global Best Bug} bucket (28.8\% of SA-like unique codes; Appendix~\ref{app:base_code_audit}), visualized in Figure~\ref{fig:logic_gap}:}
\begin{center}
\small \texttt{if new > \textbf{best} or rand < exp((new - \textbf{best})/T):}
\end{center}
\rev{By comparing neighbors to the running global best found so far rather than the current state, the Base Model's heuristics effectively ``freeze'' the search once it reaches a local maximum under the current neighborhood. Hero corrects this to $\Delta = S_{\text{new}} - S_{\text{curr}}$, enabling true annealing. The appendix audit shows that this bug is important but not exhaustive: the remaining SA-like codes are still dominated by missing best-solution tracking, ambiguous acceptance logic, and weak neighborhood structure rather than by a single isolated mistake. This is the central semantic-repair result of the paper: on SDS, RL does not just retrieve a solver family more often; it repairs the executable logic that makes that family actually work in a deceptive landscape.}

\paragraph{Code Implementation Analysis.} \rev{The generated code (Figure~\ref{fig:emergent_code} in Appendix~\ref{app:qualitative}) consistently implements three ingredients that explain the behavioral jump: (1) a \textbf{Constraint Guard} (using \texttt{while not feasible()} or conditional rejection) to prevent wasted evaluations; (2) a correct \textbf{Metropolis Criterion} for probabilistic acceptance ($P = \exp(\Delta S / T)$); and (3) \textbf{Generalization} via standard streams, decoupling the compiled solver from a single fixed problem instance. Together, these support the interpretation that RL produces a reusable solver artifact rather than a one-off answer generator.}

\section{Discussion}

\paragraph{Synthesizing Reasoning Behavior from Standard Weights.}
\rev{A critical finding is that our policy---initialized from a standard coder-instruct model (Qwen 2.5-Coder) rather than a dedicated reasoning model---outperforms baselines that rely on stronger proprietary search models at deployment time. The refreshed ShinkaEvolve baseline uses stronger mutation models, yet without the same feasibility-first alignment these deployment-time search systems still underperform on deceptive SDS instances (14.4\% gap under the fairer rerun). We therefore interpret Hero not as evidence that frontier models lack raw capability, but as evidence that a comparatively small coder model can be pushed into a distinct behavioral regime by RL and verification scaffolds.}

\paragraph{Breaking the Greedy Attractor.}
\rev{Our ablation analysis reveals that the primary driver of solver synthesis is not rich positive guidance, but control over the base model's easiest failure mode. The ``w/o Structure'' ablation confirms that without the lazy-sort penalty, the model settles into a local optimum of greedy construction ($34.7\%$ gap). By blocking this path of least resistance, the structural scaffold forces the policy to escape the greedy basin and converge on a more robust solver family. This suggests that in RL for Code, negative constraints on bad executable behavior can be more important than dense hints toward good outcomes.}

\paragraph{Scaffolding as a Contextual Key.}
\rev{Our ablation analysis (Table~\ref{tab:main_results}) also confirms that the scaffold \textit{alone} is insufficient. The Base 14B model, even when provided with the identical Hero prompt, still saturates at a 28.7\% gap. The prompt is therefore not the solution by itself; it is a contextual key whose intended procedure must still be operationalized by training. This is another way to see the competence wall: the instructions name a useful mode of behavior, but RL is what makes that mode executable on SDS.}

\paragraph{Amortized Intelligence and Verification.}
\rev{Recent frameworks like {ShinkaEvolve} \citep{lange_shinkaevolve_2025} and {ThetaEvolve} \citep{wang_thetaevolve_2025} rely on intensive test-time search. In contrast, our approach acts as a \textit{compiler}, paying the reasoning cost once during synthesis. The generated code adheres to a Simulated Annealing template ($99.8\%$ convergence) and operates on dynamic inputs, enabling a reusable deployment object instead of a fresh search loop per instance. We now directly validate that framing with a strict compile-once evaluation: extracting one fixed Hero-generated solver per seed and executing it unchanged across the full SDS test set yields 97.87\% pass rate with 4.34\% mean gap to VBS, compared with 97.83\% pass rate and 4.82\% mean gap to VBS for the standard per-instance Hero evaluation. On SDS, the learned process does produce a reusable solver rather than only a per-instance generation policy.}

\paragraph{Stronger Non-Neural Baseline.}
\rev{We also added a manually specified, constraint-aware Simulated Annealing baseline evaluated through the same fixed-code path. It reaches 95.63\% pass rate with 5.32\% mean gap to VBS across the three SDS seeds, but the frozen Hero solver still outperforms it on both feasibility and solution quality (97.87\%, 4.34\%). This shows that Hero remains competitive even against a competent manually specified heuristic from the same algorithmic family.}

\paragraph{Breaking the Sampling Ceiling.} Our results offer a critical nuance to the ``sampling hypothesis'' \citep{yue_does_2025}. While we confirm that the algorithmic primitives exist within the base distribution (e.g., 21.9\% SA-like template retrieval in the canonical raw-code audit), naive sampling fails to assemble them into a functional solver. The Base Model's failure is semantic: it retrieves the correct syntax but hallucinates the acceptance logic (Global Best Bug). Extensive sampling ($N=64$) merely yields variations of this broken physics. RL acts as a \textit{functional compiler}, bridging the gap between latent capability and operational validity by correcting the algorithm's mechanics (aligning constraint guards and Metropolis criteria). \rev{In our current experiments, RL is the method that closes this gap relative to naive sampling; evaluating other repair-oriented procedures remains an important direction for future work.}

\paragraph{Contextual Specialization.}
The negligible catastrophic forgetting observed in Section~\ref{sec:robustness} suggests that our scaffolded training protocol acts as a specific mode-switch rather than a destructive overwrite. Because the training utilized a distinct, scaffolded system prompt (``...expert engineering reasoning engine...''), the policy likely learned to associate the specialized search behaviors (Simulated Annealing) strictly with this context. When evaluated on general benchmarks (HumanEval) using standard prompts, the model gracefully reverts to its pre-trained distribution, offering a ``best of both worlds'' capability profile.

\paragraph{Process Constraints vs. Outcome Shaping.}
\rev{Our results clarify a critical distinction between \textit{outcome shaping} (e.g., oracles) and \textit{process verification} (e.g., $R_{\text{exec}}$). While we discard oracle signals to reduce reward hacking around the metric, we retain the structural scaffold as a necessary \textit{negative constraint} on bad executable behavior. The failure of the ``w/o Structure'' ablation demonstrates that simply removing the oracle is insufficient; the objective must actively devalue the base model's pre-trained bias toward linear approximations. In this sense, the scaffold is less a teacher of the final algorithm than a filter that prevents the policy from staying in the greedy basin.}

\paragraph{Feasibility Density During Training.}
\rev{We also instrumented fresh Hero-style SDS training to measure feasibility density at the true GRPO group level. Across 1,284 reconstructed groups of 64 completions, 85.83\% already contain at least one feasible sample, and the average feasible count rises from 23.32 / 64 early in training to 50.94 / 64 late in training. On SDS, the hard nominal gate therefore does not mainly create an all-zero-group regime.}

\paragraph{Normalization Sensitivity.}
\rev{A further limitation is that the SDS recipe is sensitive to the nominal reward normalization heuristic. In a matched 3-seed ablation that changes only the interaction-normalization estimate, the treatment aggregate drops from Hero's 97.83\% $\pm$ 0.15\% pass rate and 4.08\% $\pm$ 1.31\% mean gap to VBS to 70.40\% $\pm$ 24.36\% pass rate and 31.07\% $\pm$ 22.83\% mean gap to VBS. We therefore view the current normalization rule as a consequential part of the method design.}

\paragraph{Limitations.}
\rev{Important limitations remain. SDS is still a custom benchmark/testbed, so the strongest conclusions remain SDS-specific even after the JSSP validation. The recipe is not domain-agnostic without adaptation, and the SDS setup also depends materially on benchmark-specific design choices, especially the nominal reward normalization heuristic. We therefore view the paper as evidence for a disciplined solver-synthesis recipe under explicit scaffolding, not as a claim of universal robustness across optimization domains or reward parameterizations.}

\section{Conclusion}

\rev{We presented an empirical study of when RL can move a code LLM from per-instance search toward reusable solver synthesis. On SDS, naive sampling encounters a clear competence wall even when the base model retrieves superficially appropriate templates. RL closes that gap not only by concentrating on a better solver family, but by repairing the executable logic required for that family to function in a deceptive landscape. The resulting policy produces a reusable solver artifact that remains highly competitive under a strict compile-once evaluation, while additional JSSP experiments provide narrower evidence that the scaffold extends beyond SDS. At the same time, our negative ablations and limitations make clear that this is not yet a domain-agnostic recipe: the current method still depends on explicit feasibility-first scaffolding and consequential design choices such as reward normalization.}


\section*{Acknowledgements}
This work was supported by a grant from the Swiss National Supercomputing Centre (CSCS) under project ID a0105 on Alps.

\section*{Software and Data}
\rev{Code, datasets, checkpoints, and experiment artifacts are provided through the linked public repositories associated with this preprint. Appendix \ref{app:implementation} lists the dependency versions, ablation configurations, and numerical non-determinism notes needed to understand the experiments.}

\section*{Impact Statement}
This work introduces a methodology for synthesizing standalone combinatorial solvers using LLMs. While we reduce the computational cost of reasoning, we acknowledge that automated solver synthesis carries risks if deployed in safety-critical infrastructure without formal verification. The achieved 5.0\% gap to VBS indicates these solvers are not yet exact. Users should monitor for dual-use risks in domains involving resource disruption. We have mitigated the environmental footprint of our research by utilizing a parsimonious training regime (4 hours on 12 GPUs).

\bibliographystyle{unsrtnat}
\bibliography{references}

@misc{austin_program_2021,
 author = {Austin, Jacob and Odena, Augustus and Nye, Maxwell and Bosma, Maarten and Michalewski, Henryk and Dohan, David and Jiang, Ellen and Cai, Carrie and Terry, Michael and Le, Quoc and others},
 archivePrefix = {arXiv},
 doi = {10.48550/arXiv.2108.07732},
 eprint = {2108.07732},
 publisher = {arXiv},
 title = {Program {Synthesis} with {Large} {Language} {Models}},
 url = {https://arxiv.org/abs/2108.07732},
 year = {2021}
}

@misc{chen_evaluating_2021,
 author = {Chen, Mark and Tworek, Jerry and Jun, Heewoo and Yuan, Qiming and Pinto, Henrique Ponde de Oliveira and Kaplan, Jared and Edwards, Harri and Burda, Yuri and Joseph, Nicholas and Brockman, Greg and others},
 archivePrefix = {arXiv},
 doi = {10.48550/arXiv.2107.03374},
 eprint = {2107.03374},
 publisher = {arXiv},
 title = {Evaluating {Large} {Language} {Models} {Trained} on {Code}},
 url = {https://arxiv.org/abs/2107.03374},
 year = {2021}
}

@inproceedings{gao_pal_2023,
 author = {Gao, Luyu and Madaan, Aman and Zhou, Shuyan and Alon, Uri and Liu, Pengfei and Yang, Yiming and Callan, Jamie and Neubig, Graham},
 booktitle = {Proceedings of the 40th {International} {Conference} on {Machine} {Learning}},
 pages = {10764--10799},
 publisher = {PMLR},
 title = {{PAL}: {Program}-aided {Language} {Models}},
 url = {https://proceedings.mlr.press/v202/gao23f.html},
 year = {2023}
}

@misc{gehring_rlef_2025,
 author = {Gehring, Jonas and Zheng, Kunhao and Copet, Jade and Mella, Vegard and Carbonneaux, Quentin and Cohen, Taco and Synnaeve, Gabriel},
 archivePrefix = {arXiv},
 doi = {10.48550/arXiv.2410.02089},
 eprint = {2410.02089},
 publisher = {arXiv},
 title = {{RLEF}: {Grounding} {Code} {LLMs} in {Execution} {Feedback} with {Reinforcement} {Learning}},
 url = {https://arxiv.org/abs/2410.02089},
 year = {2025}
}

@misc{huang_large_2024,
 author = {Huang, Jie and Chen, Xinyun and Mishra, Swaroop and Zheng, Huaixiu Steven and Yu, Adams Wei and Song, Xinying and Zhou, Denny},
 archivePrefix = {arXiv},
 doi = {10.48550/arXiv.2310.01798},
 eprint = {2310.01798},
 publisher = {arXiv},
 title = {Large {Language} {Models} {Cannot} {Self}-{Correct} {Reasoning} {Yet}},
 url = {https://arxiv.org/abs/2310.01798},
 year = {2024}
}

@inproceedings{le_coderl_2022,
 author = {Le, Hung and Wang, Yue and Gotmare, Akhilesh Deepak and Savarese, Silvio and Hoi, Steven C. H.},
 booktitle = {Advances in {Neural} {Information} {Processing} {Systems}},
 title = {{CodeRL}: {Mastering} {Code} {Generation} through {Pretrained} {Models} and {Deep} {Reinforcement} {Learning}},
 url = {https://openreview.net/forum?id=WaGvb7OzySA},
 year = {2022}
}

@misc{hui_qwen25-coder_2024,
 author = {Hui, Binyuan and Yang, Jian and Cui, Zeyu and Yang, Jiaxi and Liu, Dayiheng and Zhang, Lei and Liu, Tianyu and Zhang, Jiajun and Yu, Bowen and Lu, Keming and others},
 archivePrefix = {arXiv},
 doi = {10.48550/arXiv.2409.12186},
 eprint = {2409.12186},
 publisher = {arXiv},
 title = {Qwen2.5-{Coder} {Technical} {Report}},
 url = {https://arxiv.org/abs/2409.12186},
 year = {2024}
}

@incollection{karp_reducibility_2010,
 author = {Karp, Richard M.},
 booktitle = {50 {Years} of {Integer} {Programming} 1958-2008: {From} the {Early} {Years} to the {State}-of-the-{Art}},
 doi = {10.1007/978-3-540-68279-0_8},
 pages = {219--241},
 publisher = {Springer},
 title = {Reducibility {Among} {Combinatorial} {Problems}},
 year = {2010}
}

@inproceedings{kool_attention_2018,
 author = {Kool, Wouter and Hoof, Herke van and Welling, Max},
 booktitle = {International Conference on Learning Representations},
 title = {Attention, {Learn} to {Solve} {Routing} {Problems}!},
 url = {https://openreview.net/forum?id=ByxBFsRqYm},
 year = {2018}
}

@misc{lange_shinkaevolve_2025,
 author = {Lange, Robert Tjarko and Imajuku, Yuki and Cetin, Edoardo},
 archivePrefix = {arXiv},
 doi = {10.48550/arXiv.2509.19349},
 eprint = {2509.19349},
 publisher = {arXiv},
 title = {{ShinkaEvolve}: {Towards} {Open}-{Ended} {And} {Sample}-{Efficient} {Program} {Evolution}},
 url = {https://arxiv.org/abs/2509.19349},
 year = {2025}
}

@misc{li_autotriton_2025,
 author = {Li, Shangzhan and Wang, Zefan and He, Ye and Li, Yuxuan and Shi, Qi and Li, Jianling and Hu, Yonggang and Che, Wanxiang and Han, Xu and Liu, Zhiyuan and others},
 archivePrefix = {arXiv},
 doi = {10.48550/arXiv.2507.05687},
 eprint = {2507.05687},
 publisher = {arXiv},
 title = {{AutoTriton}: {Automatic} {Triton} {Programming} with {Reinforcement} {Learning} in {LLMs}},
 url = {https://arxiv.org/abs/2507.05687},
 year = {2025}
}

@misc{liu_rltf_2023,
 author = {Liu, Jiate and Zhu, Yiqin and Xiao, Kaiwen and Fu, Qiang and Han, Xiao and Yang, Wei and Ye, Deheng},
 archivePrefix = {arXiv},
 doi = {10.48550/arXiv.2307.04349},
 eprint = {2307.04349},
 publisher = {arXiv},
 title = {{RLTF}: {Reinforcement} {Learning} from {Unit} {Test} {Feedback}},
 url = {https://arxiv.org/abs/2307.04349},
 year = {2023}
}

@inproceedings{manchanda_gcomb_2020,
 author = {Manchanda, Sahil and Mittal, Akash and Dhawan, Anuj and Medya, Sourav and Ranu, Sayan and Singh, Ambuj},
 booktitle = {Advances in {Neural} {Information} {Processing} {Systems}},
 pages = {20000--20011},
 publisher = {Curran Associates, Inc.},
 title = {{GCOMB}: {Learning} {Budget}-constrained {Combinatorial} {Algorithms} over {Billion}-sized {Graphs}},
 url = {https://papers.nips.cc/paper/2020/hash/e7532dbeff7ef901f2e70daacb3f452d-Abstract.html},
 volume = {33},
 year = {2020}
}

@misc{mistral-ai_magistral_2025,
 author = {Mistral-AI and Rastogi, Abhinav and Jiang, Albert Q. and Lo, Andy and Berrada, Gabrielle and Lample, Guillaume and Rute, Jason and Barmentlo, Joep and Yadav, Karmesh and Khandelwal, Kartik and others},
 archivePrefix = {arXiv},
 doi = {10.48550/arXiv.2506.10910},
 eprint = {2506.10910},
 publisher = {arXiv},
 title = {Magistral},
 url = {https://arxiv.org/abs/2506.10910},
 year = {2025}
}

@misc{novikov_alphaevolve_2025,
 author = {Novikov, Alexander and Vũ, Ngân and Eisenberger, Marvin and Dupont, Emilien and Huang, Po-Sen and Wagner, Adam Zsolt and Shirobokov, Sergey and Kozlovskii, Borislav and Ruiz, Francisco J. R. and Mehrabian, Abbas and others},
 archivePrefix = {arXiv},
 doi = {10.48550/arXiv.2506.13131},
 eprint = {2506.13131},
 publisher = {arXiv},
 title = {{AlphaEvolve}: {A} coding agent for scientific and algorithmic discovery},
 url = {https://arxiv.org/abs/2506.13131},
 year = {2025}
}

@inproceedings{pan_logic-lm_2023,
 author = {Pan, Liangming and Albalak, Alon and Wang, Xinyi and Wang, William Yang},
 booktitle = {The 2023 Conference on Empirical Methods in Natural Language Processing},
 title = {Logic-{LM}: {Empowering} {Large} {Language} {Models} with {Symbolic} {Solvers} for {Faithful} {Logical} {Reasoning}},
 url = {https://openreview.net/forum?id=nWXMv949ZH&noteId=qt0t8SsVvT},
 year = {2023}
}

@inproceedings{perron_cp-sat-lp_2023,
 author = {Perron, Laurent and Didier, Frédéric and Gay, Steven},
 booktitle = {29th {International} {Conference} on {Principles} and {Practice} of {Constraint} {Programming} ({CP} 2023)},
 doi = {10.4230/LIPIcs.CP.2023.3},
 pages = {3:1--3:2},
 publisher = {Schloss Dagstuhl – Leibniz-Zentrum für Informatik},
 title = {The {CP}-{SAT}-{LP} {Solver}},
 volume = {280},
 year = {2023}
}

@article{pisinger_quadratic_2007,
 author = {Pisinger, David},
 doi = {10.1016/j.dam.2006.08.007},
 journal = {Discrete Applied Mathematics},
 number = {5},
 pages = {623--648},
 title = {The quadratic knapsack problem—a survey},
 volume = {155},
 year = {2007}
}

@article{romera-paredes_mathematical_2024,
 author = {Romera-Paredes, Bernardino and Barekatain, Mohammadamin and Novikov, Alexander and Balog, Matej and Kumar, M. Pawan and Dupont, Emilien and Ruiz, Francisco J. R. and Ellenberg, Jordan S. and Wang, Pengming and Fawzi, Omar and others},
 doi = {10.1038/s41586-023-06924-6},
 journal = {Nature},
 number = {7995},
 pages = {468--475},
 title = {Mathematical discoveries from program search with large language models},
 volume = {625},
 year = {2024}
}

@misc{shao_deepseekmath_2024,
 author = {Shao, Zhihong and Wang, Peiyi and Zhu, Qihao and Xu, Runxin and Song, Junxiao and Bi, Xiao and Zhang, Haowei and Zhang, Mingchuan and Li, Y. K. and Wu, Y. and others},
 archivePrefix = {arXiv},
 doi = {10.48550/arXiv.2402.03300},
 eprint = {2402.03300},
 publisher = {arXiv},
 title = {{DeepSeekMath}: {Pushing} the {Limits} of {Mathematical} {Reasoning} in {Open} {Language} {Models}},
 url = {https://arxiv.org/abs/2402.03300},
 year = {2024}
}

@inproceedings{ye_reevo_2024,
 author = {Ye, Haoran and Wang, Jiarui and Cao, Zhiguang and Berto, Federico and Hua, Chuanbo and Kim, Haeyeon and Park, Jinkyoo and Song, Guojie},
 booktitle = {Advances in {Neural} {Information} {Processing} {Systems}},
 title = {{ReEvo}: {Large} {Language} {Models} as {Hyper}-{Heuristics} with {Reflective} {Evolution}},
 url = {https://openreview.net/forum?id=483IPG0HWL},
 year = {2024}
}

@article{valmeekam_planbench_2023,
 author = {Valmeekam, Karthik and Marquez, Matthew and Olmo, Alberto and Sreedharan, Sarath and Kambhampati, Subbarao},
 journal = {Advances in Neural Information Processing Systems},
 pages = {38975--38987},
 title = {{PlanBench}: {An} {Extensible} {Benchmark} for {Evaluating} {Large} {Language} {Models} on {Planning} and {Reasoning} about {Change}},
 url = {https://proceedings.neurips.cc/paper_files/paper/2023/hash/7a92bcdede88c7afd108072faf5485c8-Abstract-Datasets_and_Benchmarks.html},
 volume = {36},
 year = {2023}
}

@misc{martinasso_alps_2025,
 author = {Martinasso, Maxime and Klein, Mark and Schulthess, Thomas C.},
 archivePrefix = {arXiv},
 doi = {10.48550/arXiv.2507.02404},
 eprint = {2507.02404},
 primaryclass = {cs.DC},
 title = {Alps, a versatile research infrastructure},
 url = {https://arxiv.org/abs/2507.02404},
 year = {2025}
}

@article{alam_versatile_2023,
 author = {Alam, Sadaf R. and Gila, Miguel and Klein, Mark and Martinasso, Maxime and Schulthess, Thomas C.},
 doi = {10.1177/10943420231167811},
 journal = {The International Journal of High Performance Computing Applications},
 number = {3--4},
 pages = {288--305},
 title = {Versatile software-defined {HPC} and cloud clusters on {Alps} supercomputer for diverse workflows},
 volume = {37},
 year = {2023}
}

@misc{wang_thetaevolve_2025,
 author = {Wang, Yiping and Su, Shao-Rong and Zeng, Zhiyuan and Xu, Eva and Ren, Liliang and Yang, Xinyu and Huang, Zeyi and He, Xuehai and Ma, Luyao and Peng, Baolin and others},
 archivePrefix = {arXiv},
 doi = {10.48550/arXiv.2511.23473},
 eprint = {2511.23473},
 publisher = {arXiv},
 title = {{ThetaEvolve}: {Test}-time {Learning} on {Open} {Problems}},
 url = {https://arxiv.org/abs/2511.23473},
 year = {2025}
}

@inproceedings{yu_dapo_2025,
 author = {Yu, Qiying and Zhang, Zheng and Zhu, Ruofei and Yuan, Yufeng and Zuo, Xiaochen and Yue, Yu and Dai, Weinan and Fan, Tiantian and Liu, Gaohong and Liu, Juncai and others},
 booktitle = {The Thirty-ninth Annual Conference on Neural Information Processing Systems},
 title = {{DAPO}: {An} {Open}-{Source} {LLM} {Reinforcement} {Learning} {System} at {Scale}},
 url = {https://openreview.net/forum?id=2a36EMSSTp},
 year = {2025}
}

@misc{yue_does_2025,
 author = {Yue, Yang and Chen, Zhiqi and Lu, Rui and Zhao, Andrew and Wang, Zhaokai and Yue, Yang and Song, Shiji and Huang, Gao},
 archivePrefix = {arXiv},
 doi = {10.48550/arXiv.2504.13837},
 eprint = {2504.13837},
 publisher = {arXiv},
 title = {Does {Reinforcement} {Learning} {Really} {Incentivize} {Reasoning} {Capacity} in {LLMs} {Beyond} the {Base} {Model}?},
 url = {https://arxiv.org/abs/2504.13837},
 year = {2025}
}

@misc{openai_gpt5_2025,
 author = {Singh, Aaditya and Fry, Adam and Perelman, Adam and Tart, Adam and Ganesh, Adi and El-Kishky, Ahmed and McLaughlin, Aidan and Low, Aiden and Ostrow, AJ and others},
 archivePrefix = {arXiv},
 doi = {10.48550/arXiv.2601.03267},
 eprint = {2601.03267},
 primaryClass = {cs.CL},
 title = {{OpenAI} {GPT}-5 {System} {Card}},
 url = {https://arxiv.org/abs/2601.03267},
 year = {2025}
}

@misc{openr1,
 author = {{Hugging Face}},
 month = {January},
 title = {{Open R1}: A fully open reproduction of {DeepSeek-R1}},
 url = {https://github.com/huggingface/open-r1},
 year = {2025}
}

\newpage
\appendix

\section{System Prompt Analysis}
\label{app:prompts}

To understand the mechanism behind the model's performance, we compare the ``Hero'' system prompt (which provides algorithmic scaffolding) against the ``Ablated'' prompt (which provides only task definitions).

\subsection{The Necessity of Scaffolding}
As shown in the main text, removing the detailed prompt causes the pass rate to drop from 97.8\% to 64.4\% (Table~\ref{tab:main_results}). This significant degradation suggests that the base model (Qwen2.5-Coder) benefits heavily from \textbf{explicit meta-cognitive instructions}.

As shown in Box C.1, the Hero prompt does not tell the model \textit{which} algorithm to use (e.g., it never mentions ``Simulated Annealing''). Instead, it enforces a \textbf{reasoning structure}:
\begin{enumerate}
    \item \textbf{Deconstruct:} Force acknowledgment of the landscape's deceptiveness.
    \item \textbf{Hypothesize:} Explicitly ask for a search strategy that balances exploration/exploitation.
    \item \textbf{Critique:} Pre-emptively flag greedy approaches as failure modes.
\end{enumerate}

In contrast, the Ablated prompt (Box C.2) treats the task as a standard code generation problem. Without the ``Critique'' step, the model often defaults to greedy construction, which fails on the deceptive test set.

\begin{figure}[t!]
\centering
\begin{minipage}[t]{0.48\textwidth}
\begin{tcolorbox}[title=\textbf{Box C.1: Hero System Prompt (Scaffolded)}, colback=green!5!white, colframe=green!50!black, fonttitle=\bfseries]
\scriptsize
\begin{verbatim}
You are an expert engineering reasoning engine
specialized in combinatorial optimization. Your 
task is to write a high-performance Python script
to solve Synergistic Dependency Selection (SDS) 
problems.

FORMATTING RULES:
1. Begin with a <think> block.
2. End with a single <code> block containing the 
   JSON-processing Python script.
3. No other text is allowed outside these blocks.

THINKING GUIDELINES:
Inside the <think> block, you must engage in a 
rigorous algorithm design process:
- **Deconstruct**: Analyze the objective landscape.
  Acknowledge that simple heuristics (like greedy
  selection) will likely get stuck in local optima
  due to negative interaction weights and complex
  constraints.
- **Hypothesize**: Propose a search strategy 
  capable of exploring the solution space 
  effectively. Consider how to balance 
  "exploitation" (improving a good solution) 
  with "exploration" (escaping bad local optima).
- **Critique**: Question your approach. "Does my 
  algorithm just pick the next best item? If so, 
  it will fail on deceptive landscapes. How do I 
  add lookahead, backtracking, or iterative 
  improvement?"
- **Simulate**: Mentally dry-run the logic to 
  ensure constraints (mutex, precedence) are 
  strictly satisfied during the search.
- **Finalize**: Verify I/O requirements.

GOAL:
Your code must aim for **Global Optimality**, 
while being feasible. You must write a 
self-contained solver (no external black-box 
libraries like OR-Tools) that intelligently 
searches for the best possible score under 
constraints.
\end{verbatim}
\end{tcolorbox}
\end{minipage}%
\hfill
\begin{minipage}[t]{0.48\textwidth}
\begin{tcolorbox}[title=\textbf{Box C.2: Ablated System Prompt (Minimal)}, colback=red!5!white, colframe=red!50!black, fonttitle=\bfseries]
\scriptsize
\begin{verbatim}
You are a coding assistant. Write a Python script 
to solve the Synergistic Dependency Selection (SDS)
optimization problem.

FORMATTING RULES:
1. Begin with a <think> block.
2. End with a single <code> block containing the 
   JSON-processing Python script.
3. No other text is allowed outside these blocks.

IO CONTRACT:
- Read JSON from stdin with keys 
  {"requirements": {...}, "catalog": {...}}.
- Print JSON to stdout with key 
  {"selection": {"variables": [...]}}.
- Use the python `json` library.
\end{verbatim}
\end{tcolorbox}
\end{minipage}
\caption{\textbf{System Prompt Comparison.} The Hero prompt (left) enforces a scaffolded reasoning loop whose core is ``Deconstruct-Hypothesize-Critique'', which we find essential for escaping local optima. The Ablated prompt (right) relies solely on the model's pre-trained priors, leading to greedy failures.}
\label{fig:prompt_comparison}
\end{figure}

\section{Reward Function Details}
\label{app:rewards}

\subsection{Complete Mathematical Formulation}

The composite reward function $R$ for the Hero configuration is defined as:
\begin{equation}
R = 0.10 \cdot R_{\text{format}} + 0.20 \cdot R_{\text{exec}} + 0.70 \cdot R_{\text{nom}}
\label{eq:composite_reward}
\end{equation}

\rev{Here $R_{\text{format}}$, $R_{\text{exec}}$, and $R_{\text{nom}}$ denote the raw reward-function outputs passed to GRPO, while the coefficients in Eq.~\ref{eq:composite_reward} are the trainer-level weights specified in the YAML configuration. In particular, $R_{\text{exec}}$ is an intentionally shaped, unnormalized execution reward composed of syntax, schema, structure, and feasibility terms; its internal subcomponent weights are therefore implementation-level shaping choices, not a second normalized decomposition of the final scalar reward.}

where $R_{\text{nom}}$ is the nominal reward (normalized score on the prompt's mission). Note that the Hero configuration explicitly excludes oracle anchoring to minimize variance.

\subsubsection{Format Reward ($R_{\text{format}}$)}
The format reward is a binary indicator enforcing a strict output structure:
\begin{equation}
R_{\text{format}} = \mathbb{I}(\text{exactly one } \langle\text{think}\rangle \text{ block}) \\
\cdot \mathbb{I}(\text{exactly one } \langle\text{code}\rangle \text{ block}) \cdot \mathbb{I}(\text{correct ordering})
\end{equation}

\subsubsection{Code Execution Reward ($R_{\text{exec}}$)}
\label{app:exec_reward_defs}

We decompose the execution-validity reward into four sub-components:
\begin{equation}
R_{\text{exec}} = R_{\text{syntax}} + R_{\text{schema}} + R_{\text{structure}} + R_{\text{feasibility}}.
\end{equation}

\paragraph{Indicator Definitions.}
We define the indicator events used in Eq.~(10) (main text) as follows.

\begin{itemize}
    \item \textbf{Syntax Valid} ($\mathbb{I}(\text{syntax valid})$): after extracting the \texttt{<code>} block and passing a lightweight structural validator, the code is executed in a sandbox with a fixed timeout. We set
    \[
    \mathbb{I}(\text{syntax valid}) \;=\; \mathbb{I}(\texttt{run\_candidate} \text{ returns without an } ``\texttt{error}'' \text{ field}).
    \]
    If execution fails (runtime error / timeout), $R_{\text{exec}}$ is set to $0$ for that sample.

    \item \textbf{Schema Valid} ($\mathbb{I}(\text{schema valid})$): given successful execution, we require the program to output a JSON object containing a \texttt{selection} field. Concretely,
    \[
    \mathbb{I}(\text{schema valid}) \;=\; \mathbb{I}(\texttt{``selection''} \in \texttt{result}),
    \]
    where \texttt{result} is the parsed output returned by the evaluator. (This is intentionally a minimal schema check used for dense shaping; full feasibility is enforced separately via $N_{\text{vio}}$.)

    \item \textbf{Graph-Theoretic Structure} ($\mathbb{I}(\text{graph-theoretic structure})$): a lightweight static feature detector encourages non-degenerate SDS solvers early in training. Let \texttt{code\_lower} be the lowercased generated code. We set
    \[
    \mathbb{I}(\text{graph-theoretic structure}) \;=\;
    \mathbb{I}\Big(\exists k \in \mathcal{K}_{\text{graph}} \text{ such that } k \in \texttt{code\_lower}\Big),
    \]
    where $\mathcal{K}_{\text{graph}}$ is a fixed keyword set defined as:
    \[
    \begin{aligned}
    \mathcal{K}_{\text{graph}}=\{&\texttt{networkx}, \texttt{adjacency}, \texttt{neighbor}, \texttt{interactions}, \texttt{precedence}, \\
    &\texttt{mutex}, \texttt{recursion}, \texttt{memoization}, \texttt{backtrack}, \\ 
    &\texttt{graph}, \texttt{edge}, \texttt{vertex}, \texttt{topological}, \texttt{dag}\}.
    \end{aligned}
    \]
    Additionally, we apply an explicit anti-mode-collapse penalty for ``lazy sort'' solutions:
    if \texttt{``sorted''} and \texttt{``weight''} appear in \texttt{code\_lower} while \texttt{``interactions''} does \emph{not}, we subtract $0.2$ from $R_{\text{exec}}$.
\end{itemize}

\paragraph{Component Formulation.}
With these indicators, the reward components are:
\begin{align}
R_{\text{syntax}} &= 0.1 \cdot \mathbb{I}(\text{syntax valid}), \\
R_{\text{schema}} &= 0.1 \cdot \mathbb{I}(\text{schema valid}), \\
R_{\text{structure}} &= 0.2 \cdot \alpha(t)\cdot \mathbb{I}(\text{graph-theoretic structure}), \\
R_{\text{feasibility}} &= 0.3 \cdot \mathbb{I}(N_{\text{vio}}=0) \;-\; \min(0.2, 0.03 \cdot N_{\text{vio}})\cdot \mathbb{I}(N_{\text{vio}}>0).
\end{align}

\paragraph{Curriculum Fading.}
We use curriculum fading to emphasize structured, non-hallucinated code early in training:
\[
\alpha(t)=
\begin{cases}
1.0 & \text{if } t < 0.4,\\
0.2 & \text{otherwise},
\end{cases}
\]
where $t\in[0,1]$ is normalized training progress (current step divided by total steps).

\paragraph{Violation Count.}
$N_{\text{vio}}$ is the number of constraint violations computed by a lightweight checker over the SDS constraints (precedence, mutex, group constraints, and cardinality bounds) applied to the produced \texttt{selection.variables}.

\subsubsection{Nominal Reward ($R_{\text{nom}}$)}
The primary optimization signal, gated by feasibility.
\begin{equation}
R_{\text{nom}} = 
\begin{cases} 
\text{Normalize}(\text{Score}(o)) \in [0, 1] & \text{if } N_{\text{vio}} = 0 \\
0.0 & \text{if } N_{\text{vio}} > 0
\end{cases}
\end{equation}
where $\text{Normalize}(\cdot)$ maps the raw SDS objective value to $[0, 1]$ via the simulator's reward registry. This strict gating forces the model to prioritize constraint satisfaction over greedy reward hacking, creating a ``feasibility cliff'' that prevents mode collapse into invalid solution regions.

\paragraph{Normalization Heuristic.}
The normalization function $\text{Normalize}(\cdot)$ uses a robust heuristic to map raw SDS scores to $[0, 1]$ without requiring exact knowledge of the global optimum. Given a raw score $s$, weights $\mathbf{w} = \{w_1, \ldots, w_n\}$, interactions $\mathbf{I} = \{I_{ij} : (i,j) \in \mathcal{E}\}$, and cardinality bounds $[L, U]$, the normalization proceeds as follows:

\textbf{Case 1: Meaningful Objective Structure} ($\sum_i |w_i| > \epsilon$ or $\sum_{(i,j)} |I_{ij}| > \epsilon$):
\begin{enumerate}
    \item \textbf{Weight Contribution Estimate:} Let $\mathbf{w}^+ = \{w_i : w_i > 0\}$ be the positive weights, sorted in descending order. The maximum weight contribution is estimated as:
    \begin{equation}
        W_{\max} = \sum_{i=1}^{\min(|\mathbf{w}^+|, U)} w^+_i
    \end{equation}
    where $w^+_i$ denotes the $i$-th largest positive weight. This assumes we can select up to $U$ variables with the highest positive weights.
    
    \item \textbf{Interaction Contribution Estimate:} Let $\mathbf{I}^+ = \{I_{ij} : I_{ij} > 0\}$ be the positive interactions. With $U$ selected variables, at most $\binom{U}{2} = U(U-1)/2$ pairs can contribute. The maximum interaction contribution is estimated as:
    \begin{equation}
        I_{\max} = \bar{I}^+ \cdot \min\left(|\mathbf{I}^+|, \binom{U}{2}\right)
    \end{equation}
    where $\bar{I}^+ = \frac{1}{|\mathbf{I}^+|}\sum_{I_{ij} \in \mathbf{I}^+} I_{ij}$ is the average positive interaction value when $|\mathbf{I}^+| > 0$, and $\bar{I}^+ = 0$ when $|\mathbf{I}^+| = 0$. This heuristic assumes we can achieve roughly the average positive interaction for each feasible pair.
    
    \item \textbf{Normalization:} The normalization baseline is $B = W_{\max} + I_{\max}$. If $B > \epsilon$, the normalized score is:
    \begin{equation}
        \text{Normalize}(s) = \frac{s}{B}
    \end{equation}
    Otherwise, we fall back to absolute normalization: $\text{Normalize}(s) = s / \max(\sum_i |w_i| + \sum_{(i,j)} |I_{ij}|, 1)$.
\end{enumerate}

\textbf{Case 2: Degenerate Objective} ($\sum_i |w_i| \leq \epsilon$ and $\sum_{(i,j)} |I_{ij}| \leq \epsilon$):
When the problem has no clear objective structure, we use magnitude-based normalization:
\begin{equation}
    \text{Normalize}(s) = \begin{cases}
        0.5 \cdot \left(1 + \frac{s}{1 + |s|}\right) & \text{if } |s| > \epsilon \\
        0.5 & \text{if } |s| \leq \epsilon
    \end{cases}
\end{equation}
This sigmoid-like transformation maps large positive scores to $\approx 1.0$, large negative scores to $\approx 0.0$, and zero scores to $0.5$, without requiring a maximum estimate.

\textbf{Intermediate Score Processing:} After normalization, we compute an intermediate score by clamping to $[0, 1]$ and applying a constraint penalty:
\begin{equation}
    \widetilde{R}_{\text{score}} = \max\left(0, \min\left(1, \text{Normalize}(s) - \min(N_{\text{vio}}, 1.0)\right)\right)
\end{equation}
where $N_{\text{vio}}$ is the number of constraint violations. This intermediate score remains in $[0, 1]$, but the authoritative definition of $R_{\text{nom}}$ is still the hard feasibility gate in the main methodology: if $N_{\text{vio}} > 0$, then $R_{\text{nom}} = 0$ regardless of $\widetilde{R}_{\text{score}}$.

This heuristic provides stable normalization across diverse problem structures without requiring expensive global optimization to determine exact maxima, enabling efficient reward computation during training.

\subsection{Rationale for Reward Composition}

\paragraph{Hierarchical Weight Selection.}
We selected the composite weighting $\lambda = (0.1, 0.2, 0.7)$ based on a principle of \textit{hierarchical stability} rather than exhaustive grid search. The weighting schema enforces a functional priority:
\begin{itemize}
    \item \textbf{Dominance of Optimality ($0.7$):} The nominal reward $R_{\text{nom}}$ is set as the dominant term to ensure that valid-but-suboptimal solutions cannot outscore invalid-but-promising candidates purely on auxiliary signals. This prevents the policy from ``gaming'' the auxiliary rewards.
    \item \textbf{Structural Bootstrapping ($0.2$):} The execution reward serves as a non-dominant bootstrapping signal. It provides just enough gradient mass to lift the policy out of the ``syntax-only'' local optima (where the model writes valid Python that does nothing useful) without overriding the primary optimization objective.
    \item \textbf{Format Compliance ($0.1$):} A minimal signal is sufficient to maintain the \texttt{<think>}/\texttt{<code>} structure once learned.
\end{itemize}

\paragraph{Justification of Structural Scaffolding.}
We emphasize that $R_{\text{structure}}$ acts as a \textit{semantic type-check}, not a solution leak. Specifically, the anti-mode-collapse penalty (subtracting $0.2$ if code sorts by \texttt{weight} but ignores \texttt{interactions}) is the decisive factor for overcoming the pre-trained prior, and in the implementation this penalty is applied directly within $R_{\text{exec}}$ while the positive graph-structure bonus is the faded $R_{\text{structure}}$ term. Without this negative constraint, models default to linear approximations (``lazy sorting'') which satisfy basic syntax but fail on deceptive landscapes. Our ``w/o Structure'' ablation (Table~\ref{tab:main_results}) confirms this: removing the structural signal caused the gap to VBS to degrade from $5.0\%$ to $34.7\%$, as the policy reverted to safe but suboptimal greedy heuristics. Thus, the structural reward does not teach the model \textit{how} to solve the problem (e.g., SA vs. DFS), but rather \textit{what not to do} (ignore interactions), creating the necessary pressure for advanced meta-heuristics to emerge.

\subsection{Ablations, Sensitivity, and Diagnostics}
\label{app:ablation_rewards}

\rev{To validate our scaffolded-verification hypothesis, we separate three experiment types. First, we run \emph{configuration ablations}, which directly modify the Hero recipe and therefore belong in Table~\ref{tab:ablation_configs}. Second, we run a matched \emph{sensitivity analysis} on reward normalization, which perturbs one load-bearing Hero design choice while holding the rest of the stack fixed. Third, we run \emph{diagnostics} such as feasibility sparsity and timeout analysis, which explain why the recipe succeeds or fails but are not alternative training configurations.}
\rev{Table~\ref{tab:ablation_configs} therefore summarizes the five primary configuration ablations: +Oracle, +Diversity, +Soft Gate, w/o Structure, and w/o Prompt. The normalization and training-dynamics studies are discussed separately below because they are not parallel recipe rows.}

\begin{table}[h]
\centering
\caption{\rev{Primary configuration ablations:} Hero vs. matched recipe changes (additions and subtractions). \rev{This table excludes the separate normalization sensitivity study and training diagnostics, which are discussed below because they are not parallel recipe configurations.}}
\label{tab:ablation_configs}
\small
\begin{tabular}{lccccc}
\toprule
Configuration & Format & Exec Reward & Nominal/Gen & Diversity & System Prompt \\
\midrule
Hero (Baseline) & 0.10 & Base (0.20) & Nominal (0.70) & None & Scaffolding \\
+ Oracle & 0.10 & Base + Anchor & Nominal (0.70) & None & Scaffolding \\
+ Diversity & 0.10 & Base (0.20) & Nominal (0.60) & 0.10 & Scaffolding \\
\rev{+ Soft Gate} & \rev{0.10} & \rev{Base (0.20)} & \rev{Soft Nominal (0.70)} & \rev{None} & \rev{Scaffolding} \\
w/o Structure & 0.10 & Minimal (0.20) & Nominal (0.70) & None & Scaffolding \\
w/o Prompt & 0.10 & Base (0.20) & Nominal (0.70) & None & Minimal \\
\bottomrule
\end{tabular}
\end{table}

\subsubsection{+ Oracle Anchoring (Addition)}
The \textbf{Oracle Anchoring} ablation adds a relative performance term to the base reward. It uses a piecewise function:
\begin{equation}
R_{\text{anchoring}} = \begin{cases}
1.0 + 10.0 \cdot \delta_{\text{norm}} & \text{if } \delta_{\text{norm}} > 0.001 \text{ (beating greedy)} \\
0.0 & \text{if } |\delta_{\text{norm}}| \leq 0.001 \\
-0.5 & \text{if } \delta_{\text{norm}} < -0.001 \text{ (losing to greedy)}
\end{cases}
\end{equation}
where $\delta_{\text{norm}}$ is the normalized score difference relative to the greedy baseline. As noted in the main text, this creates a signal conflict with the feasibility-gated nominal reward, inflating variance in GRPO's standardized advantages and causing training instability.

\subsubsection{+ Diversity Penalty (Addition)}
The \textbf{Diversity} ablation adds a penalty $R_{\text{div}}$ using 4-gram Jaccard similarity to penalize low entropy within the generation group ($G=64$).

\rev{\subsubsection{+ Soft Gate (Alternative)}}
\rev{The \textbf{Soft Gate} ablation keeps Hero's format reward, execution reward, prompt, dataset, and training setup fixed, but replaces the hard nominal feasibility cliff with a graded penalty on simulator-reported violations. Concretely, it uses}
\begin{equation}
\rev{R_{\text{soft}} = \max\left(0, \min\left(1, \mathrm{Normalize}(s) - 0.15 \cdot N_{\text{vio}}\right)\right).}
\end{equation}
\rev{This ablation asks whether partial credit for mildly infeasible yet high-scoring candidates improves learning. It does not: over seeds 101/202/303, the soft-gate variant reaches a lower average pass rate than canonical Hero (57.40\% vs.\ 97.83\%) and a worse shared-VBS mean gap (43.50\% vs.\ 4.09\%). Because infeasible solutions incur 100\% gap in this comparison, the larger gap is partly a consequence of the much higher infeasibility rate; accordingly, we interpret pass rate and gap jointly. Its dominant failure mode remains constraint violations, especially precedence violations. On SDS, this suggests that the hard gate is not merely a source of harmful reward sparsity; it provides a useful feasibility-first learning signal.}

\rev{\subsubsection{Matched Sensitivity Study: Reward Normalization}}
\rev{We also ran a matched 3-seed ablation that changes only the SDS nominal reward normalization heuristic while keeping the prompt, reward-stack structure, reward weights, dataset family, training budget, and SDS evaluation pipeline fixed. The control uses the normalization described in Appendix~\ref{app:rewards}, in which the positive interaction term is estimated as the average positive interaction value times the maximum feasible number of pairs. The treatment keeps the weight term unchanged but replaces that interaction estimate with the sum of the largest feasible positive interactions, i.e.,}
\begin{equation}
\rev{I_{\max}^{\text{alt}} = \sum_{k=1}^{\min(|\mathbf{I}^+|,\binom{U}{2})} I^+_{(k)},}
\end{equation}
\rev{where $I^+_{(k)}$ denotes the $k$-th largest positive interaction. Under this ablation, canonical Hero achieves 97.83\% $\pm$ 0.15\% pass rate and 4.08\% $\pm$ 1.31\% mean gap to VBS across seeds 101/202/303, whereas the treatment reaches 70.40\% $\pm$ 24.36\% pass rate and 31.07\% $\pm$ 22.83\% mean gap to VBS. The seed-level treatment results are 57.50\% / 42.95\% (seed101), 55.20\% / 45.51\% (seed202), and 98.50\% / 4.75\% (seed303). The dominant degradation mode is feasibility and constraint satisfaction: aggregated treatment errors are none=2112, constraint=829, and timeout=59. Because our standard SDS gap metric assigns infeasible rows an effective score of zero, those rows contribute 100\% gap whenever the VBS score is positive, so the gap degradation is partly a consequence of the increased infeasibility rate. We therefore do not interpret this ablation as evidence of robustness. Instead, it indicates that the current normalization heuristic is a consequential, load-bearing design choice, and that the SDS recipe is strongly seed-sensitive to perturbations of that term.}

\rev{\subsubsection{Training Diagnostic: Feasibility Sparsity During GRPO Training}}
\rev{An important concern with a feasibility-gated nominal reward is that it could be too sparse to drive learning if a GRPO comparison group of 64 completions often contains no feasible samples at all. We first attempted to recover this statistic from historic logs and cached artifacts, but those artifacts did not preserve a stable SDS mission identity sufficient for exact reconstruction of the original 64-sample groups. We therefore instrumented fresh Hero-style SDS training and reran seeds 101 / 202 / 303 with rank-sharded logging inside \texttt{open-r1}. Each rank logged lossless feasibility shards and raw generation traces, after which we reconstructed the global 64-sample groups offline by matching on \texttt{reward\_call\_index} and stable problem identity.}

\rev{Across 1,284 reconstructed GRPO groups of size 64, 85.83\% of groups contain at least one feasible completion, the average group contains 37.94 feasible completions, and the pooled feasible completion rate is 59.27\%. In a pooled early / middle / late view, the fraction of groups with any feasible completion is 88.76\%, 82.48\%, and 86.25\%, while the mean feasible count per group rises from 23.32 / 64 to 39.48 / 64 to 50.94 / 64. Thus, the nominal signal is not sparse in the sense of an all-infeasible-group regime dominating training. The more important learning-dynamics change is that feasible completions become denser within each GRPO group over time. This does not mean every early group is already rich in feasible candidates, but it does indicate that the hard nominal gate is not simply starving GRPO of comparisons on SDS.}

\subsubsection{w/o Structure (Subtraction)}
The \textbf{w/o Structure} ablation removes the graph-theoretic structure detection and curriculum fading components from the execution reward. While the Hero configuration uses a composite execution reward that includes syntax validation (0.1), schema correctness (0.1), graph-theoretic structure detection with curriculum fading (0.2), and constraint satisfaction (0.3), the Minimalist execution reward (`minimal\_feasibility\_reward`) strips away the structural checks and curriculum mechanisms. It provides a simplified three-tier reward: $1.0$ for valid feasible solutions, $0.5$ for execution success with constraint violations, $0.1$ for syntax success with wrong schema, and $0.0$ otherwise. This ablation tests whether the structural detection and curriculum fading components are necessary for learning effective search strategies. Results show that removing these components causes pass rates to drop to $74.8 \pm 19.2\%$ (compared to Hero's $97.8 \pm 0.2\%$), with a $34.7 \pm 14.1\%$ gap to VBS, indicating that the structural guidance in the execution reward is beneficial for constraint satisfaction and optimization quality.

\section{Experimental Details}
\label{app:experimental_details}

\subsection{Hyperparameters}
We employ the \texttt{open-r1} framework \citep{openr1}, which leverages the \texttt{trl} library's implementation of GRPO. This open-source framework provides a robust training recipe which we adapted for our specific reward structure. Table~\ref{tab:grpo_hyperparams} lists the complete configuration.

\begin{table}[h]
\centering
\caption{Complete GRPO hyperparameter configuration for the 14B model.}
\label{tab:grpo_hyperparams}
\small
\begin{tabular}{ll}
\toprule
Parameter & Value \\
\midrule
Base Model & Qwen2.5-Coder-14B-Instruct \\
Precision & bfloat16 (Flash Attention 2) \\
\midrule
Algorithm & GRPO \\
$\beta$ (KL Coeff) & 0.0 \\
Group Size ($G$) & 64 \\
Clip Range Lower ($\epsilon$) & 0.10 \\
Clip Range Upper ($\epsilon_{\text{high}}$) & 0.26 \citep{mistral-ai_magistral_2025} \\
Loss Type & BNPO \\
\midrule
Learning Rate & $2 \times 10^{-6}$ \\
Scheduler & Cosine (No Warmup) \\
Effective Batch Size & 4 Prompts (Limited by $G=64$ memory footprint) \\
Training GPU Configuration & 8 Training GPUs + 4 vLLM Server GPUs \\
Samples Seen & 360 Prompts ($\approx$4.5\% of full prompt-level epoch) \\
Episodes Generated & 23,040 ($360 \times 64$) \\
\midrule
Max Sequence Length & 3072 \\
Training Steps & 90 \\
Total Duration & $\approx$ 48 GPU-Hours (4h $\times$ 12 GPUs) \\
\bottomrule
\end{tabular}
\end{table}

\subsection{Dataset Details}
The training dataset consists of 10,000 instances. Table~\ref{tab:dataset_generation} details the generation parameters used to ensure landscape diversity.

\begin{table}[h]
\centering
\caption{Problem type distribution and parameters for SDS dataset generation.}
\label{tab:dataset_generation}
\small
\begin{tabular}{lcc}
\toprule
Problem Type & Weight & Key Parameters \\
\midrule
Dense Deceptive & 20\% & $w_i \in [-2, 2]$, $W_{ij} \in [-20, 20]$ \\
Structural Trap & 15\% & Chain length 4--7, bait 100.0, trap -10.0 \\
BnB Showcase & 15\% & $w_i \in [-1, 1]$, $W_{ij} \in [-25, 25]$ \\
Random SDS & 5\% & $N \in [50, 100]$ (scaling test) \\
Tree-Structured & 5\% & Weight scale 5.0, pair scale 15.0 \\
Greedy-Easy & 5\% & Designed for greedy success \\
Decomposable & 10\% & 4 disjoint clusters \\
Local Optima & 10\% & Multiple local optima \\
Planted QUBO & 10\% & Signal 5.0, noise 2.0 \\
Max-Cut QUBO & 5\% & Edge prob 0.6 \\
\bottomrule
\end{tabular}
\end{table}

\subsection{Difficulty Classification Methodology}
\label{app:difficulty}

The evaluation pipeline classifies problem instances into three difficulty categories (\textit{Trivial}, \textit{Moderate}, \textit{Hard}) based on how well a simple greedy heuristic performs relative to the best-known solution. This classification enables stratified analysis to understand how different methods perform across varying problem complexity.

\subsubsection{Virtual Best Solver (VBS) Calculation}

The \textbf{Virtual Best Solver (VBS)} represents the best score achieved by any solver (LLM or baseline) on a given problem instance. There are two levels of VBS calculation:

\textbf{Per-Experiment VBS (Individual Evaluation):} During individual evaluation runs, for each instance we collect all feasible scores from:
\begin{itemize}
    \item The LLM solution (if feasible)
    \item All baseline solvers (Greedy, Local Search, BnB, CP-SAT)
\end{itemize}

The per-experiment VBS is the maximum of these feasible scores:
\begin{equation}
\text{VBS}_{\text{exp}} = \max\{S_{\text{LLM}}, S_{\text{Greedy}}, S_{\text{Local Search}}, S_{\text{BnB}}, S_{\text{CP-SAT}}\}
\end{equation}

\textbf{Global VBS (Aggregation):} When aggregating results across multiple methods and experiments, we compute a \textbf{global VBS} per problem instance to ensure fair comparison. The global VBS is the maximum across \textit{all} methods (all LLM methods, all baselines, and the union of all $N=64$ Base model samples):
\begin{equation}
\text{VBS}_{\text{global}} = \max\{S_{\text{Method}_1}, S_{\text{Method}_2}, \ldots, S_{\text{Baseline}_1}, \ldots, S_{\text{Base}_{1..64}}\}
\end{equation}

All reported SDS gaps in aggregated results (tables and plots) use the global VBS. This prevents methods from appearing artificially optimal simply because they found a better solution than deterministic baselines but worse than the best feasible score observed anywhere in the comparison set. We therefore interpret VBS-relative gaps as gaps to the strongest observed feasible reference, not as certified distances to the true optimum.

Only feasible solutions are considered; infeasible solutions are excluded. If all solvers fail, VBS is set to $-\infty$ and the instance is classified as ``Hard.''

\subsubsection{Hardness Metric}

The \textbf{hardness} metric measures the relative gap between the greedy baseline and the VBS:

\begin{equation}
\text{hardness} = \frac{\text{VBS} - S_{\text{Greedy}}}{\max(|\text{VBS}|, \epsilon)}
\end{equation}

where $\epsilon = 10^{-10}$ prevents division by zero. Edge cases:
\begin{itemize}
    \item If all solvers fail: $\text{hardness} = 1.0$ (maximum difficulty)
    \item If greedy fails but VBS succeeds: $\text{hardness} = 1.0$ (greedy completely failed)
    \item If VBS $\leq 0$: Uses $|\text{VBS}|$ in denominator to handle negative objectives
\end{itemize}

\subsubsection{Classification Thresholds}

Instances are classified into three categories:
\begin{itemize}
    \item \textbf{Trivial}: $\text{hardness} < 0.01$ (greedy is nearly optimal, $<1\%$ gap)
    \item \textbf{Moderate}: $0.01 \leq \text{hardness} < 0.10$ ($1$--$10\%$ gap)
    \item \textbf{Hard}: $\text{hardness} \geq 0.10$ (greedy fails significantly, $\geq 10\%$ gap)
\end{itemize}

This classification is used in the stratified box plot (Figure~\ref{fig:stratified}) to visualize how methods perform across different problem difficulty levels.

\subsection{Hardware and Compute}
\begin{itemize}
    \item \textbf{Training:} 3 nodes of NVIDIA GH200 GPUs (12 GPUs total) on Alps at CSCS \citep{martinasso_alps_2025,alam_versatile_2023}.
    \item \textbf{Cost:} The final Hero model required approx. 48 GPU-hours to train.
    \item \textbf{Evaluation:} 32 CPU cores per evaluation job.
\end{itemize}

\rev{\subsection{Appendix Support for Amortized Solver Synthesis}}
\rev{\label{app:baseline_eval_additions}}

\rev{\paragraph{Fixed-code evaluation protocol.}
To directly test the compile-once claim, we extended the SDS evaluator with a fixed-code mode that reads one solver file from disk and executes it unchanged across the full held-out SDS test split. This path preserves the same scoring, VBS accounting, and output schema as the standard evaluator, including \texttt{metrics\_final.csv}, per-method tables, and experiment metadata. We use this same fixed-code path for both the frozen Hero solver and the manually specified Simulated Annealing baseline so that the comparison is matched not only in time budget but also in evaluator semantics.}

\rev{\paragraph{Frozen Hero selection rule.}
We do not hand-pick a favorable solver. For each SDS seed, we load the canonical Hero \texttt{metrics\_final.csv}, filter rows to \texttt{feasible == True} and \texttt{error\_type == "none"}, sort the surviving rows by \texttt{uuid}, and extract the first code snippet. This deterministic rule yields one frozen solver per seed, which is then executed unchanged across the corresponding 1{,}000-instance SDS test split. The resulting compile-once evaluation yields 97.87\% pass rate with 4.34\% mean gap to VBS, compared with 97.83\% pass rate and 4.82\% mean gap to VBS for standard per-instance Hero evaluation.}

\rev{\paragraph{Manually specified Simulated Annealing baseline.}
We also implemented a manually specified, constraint-aware Simulated Annealing baseline under the same fixed-code evaluator contract. The solver uses feasible initialization, neighborhood proposals that respect or repair SDS constraints, and a standard Metropolis acceptance rule against the \emph{current} state rather than the global best state. We emphasize that this is a separate manually specified implementation in the same high-level algorithm family as the converged Hero template, not a copy of a frozen Hero solver. Under the same fixed-code path, this baseline reaches 95.63\% pass rate with 5.32\% mean gap to VBS across the three SDS seeds.}

\rev{\paragraph{Runtime accounting for the baseline bundle.}
To clarify cost accounting, we produced fresh representative seed101 timing reruns for five methods: standard Hero, Frozen Hero, the manually specified Simulated Annealing baseline, Base Best-of-64, and a refreshed ShinkaEvolve baseline. For Hero and Base Best-of-64, the timing summary records both generation and evaluation wall-clock. For Frozen Hero, the manually specified Simulated Annealing baseline, and ShinkaEvolve, generation cost is zero at evaluation time and only evaluator-side execution is measured. This wall-clock accounting is separate from the main-text ``Time (s)'' column: in the main SDS figures/tables, Hero/ShinkaEvolve use post-generation solver execution cost, whereas Base Best-of-64 uses cumulative per-instance sampled-candidate execution/selection cost, still excluding LLM inference. The resulting representative wall-clock totals are 2731s for standard Hero, 2342s for Frozen Hero, 3103s for the manually specified SA baseline, 25480s for Base Best-of-64, and 2109s for refreshed ShinkaEvolve. We therefore interpret the strongest deployment contrast as one between reusable compiled solvers and cumulative multi-sample test-time search, not as a comparison of full training cost.}

\subsection{Additional Domain Beyond SDS}
\label{app:transfer_domains}

\paragraph{Why this domain.}
\rev{We use JSSP as the main additional domain because it is a standard combinatorial optimization benchmark, structurally different from SDS, and still compatible with the ``LLM writes a standalone solver'' framing. The part of the recipe we attempt to carry over is the high-level scaffold: same model family, same GRPO setup, same reward-stack structure and weights, and the same prompt philosophy of writing a white-box solver rather than emitting per-instance answers. The parts we still adapt per domain are the public solver contract, simulator/objective, native baselines, and light semantic shaping needed to enforce that contract. We therefore present JSSP as evidence that the recipe extends beyond SDS, not as evidence that no domain adaptation is needed.}

\paragraph{Job Shop Scheduling Problem (JSSP).}
\rev{JSSP consists of $J$ jobs and $M$ machines. Each job is an ordered list of operations, each operation must run on a specified machine for a fixed processing time, and each machine can process at most one operation at a time. Operations are non-preemptive and must respect within-job precedence. The objective is to minimize the makespan, i.e., the completion time of the last operation. In our text-suite evaluator, JSSP is scored against native baselines including OR-Tools, Local Search, Most Work Remaining (MWKR), Most Operations Remaining (MOPR), Shortest Processing Time (SPT), and Longest Processing Time (LPT), with solution quality reported as gap to a domain-specific VBS.}

\paragraph{Additional-domain evaluation package.}
\rev{For the additional-domain evaluation reported here, we freeze one JSSP regime and evaluate the final saved checkpoint for each of seeds 101 / 202 / 303 under the same native-baseline evaluator. This avoids per-seed checkpoint selection and keeps the beyond-SDS claim tied to one consistent recipe. All results are reported as feasibility plus gap to a JSSP-specific VBS, with native baselines OR-Tools, Local Search, Most Work Remaining (MWKR), Most Operations Remaining (MOPR), Shortest Processing Time (SPT), and Longest Processing Time (LPT).}

\paragraph{JSSP additional-domain result.}
\rev{Across the three JSSP seeds, Hero achieves a mean pass rate of $99.3\%\pm1.2\%$ and a mean gap to VBS of $6.12\%\pm3.80\%$. The seed-level results are: seed101, $100.0\%$ pass and $10.51\%$ gap; seed202, $100.0\%$ pass and $3.92\%$ gap; seed303, $98.0\%$ pass and $3.93\%$ gap. In every seed-specific local bundle, Hero outperforms the heuristic JSSP baselines and trails only OR-Tools. This is the main additional-domain result we use to support the claim that the solver-synthesis recipe extends beyond SDS.}

\begin{figure}[t]
    \centering
    \includegraphics[width=0.55\textwidth]{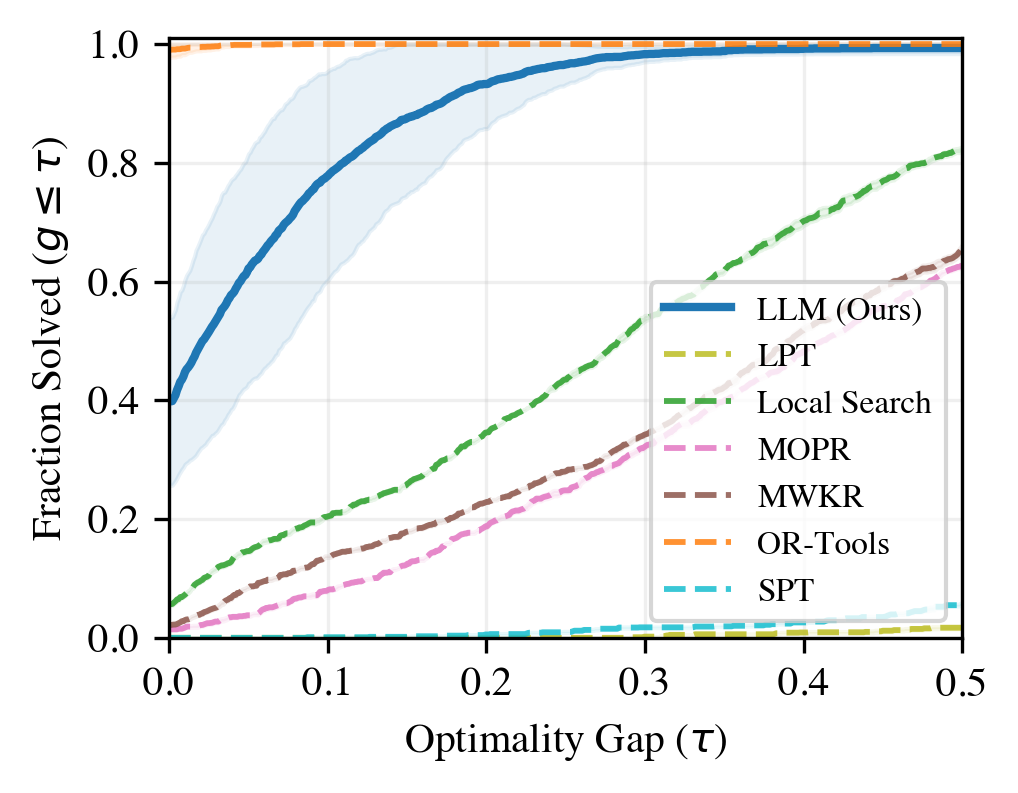}
    \caption{\rev{JSSP robustness profile aggregated across seeds 101 / 202 / 303 in the same style as Figure~\ref{fig:robustness} for SDS. For each method, the line is the mean fraction of held-out JSSP instances with gap to VBS at most $\tau$, and the shaded band shows $\pm 1$ standard deviation across seeds.}}
    \label{fig:jssp_robustness_across_seeds}
\end{figure}

\rev{Figure~\ref{fig:jssp_robustness_across_seeds} mirrors the SDS robustness-profile convention: for each method we compute one JSSP robustness curve per seed, then plot the mean coverage across seeds with a $\pm 1$ standard-deviation band. The resulting curve shows that Hero maintains the strongest low-gap coverage among the heuristic baselines across the frozen three-seed package, trailing only OR-Tools.}

\subsection{ShinkaEvolve Baseline Configuration}
\label{app:shinkaevolve}

\rev{We evaluate against ShinkaEvolve \citep{lange_shinkaevolve_2025} using 1000-sample SDS datasets generated by evolving 100 base solutions under a neutral SDS prompt and augmenting them via trace paraphrasing and problem variation. This appendix section makes the fairness story explicit: the manuscript-reported Hero and ShinkaEvolve results are evaluated on the same held-out SDS test split, with the same evaluator and the same global VBS accounting. We keep the initializer, evolution workflow, model family, and search budget fixed, and only neutralize prompt content so that the comparison tests solver-search strength rather than prompt asymmetry. The `ShinkaEvolve-SDS-1000-v2-seed\{seed\}` dataset is the Shinka solution corpus used to produce and score evolved candidates, not a distinct easier final evaluation benchmark.}

\textbf{Evolution Configuration:}
\begin{itemize}
    \item \textbf{Generations:} 10 per problem instance
    \item \textbf{LLM Models:} GPT-5-nano and GPT-5-mini (OpenAI) were used for code mutations as strong proprietary reasoning models in the deployment-time search baseline.
    \item \textbf{Patch Types:} Diff-based (60\%), full replacement (30\%), cross-over (10\%)
    \item \textbf{Island Model:} 2 parallel populations with migration every 10 generations
    \item \textbf{Archive Size:} 20 elite solutions per island
    \item \textbf{Selection:} Weighted parent selection with $\lambda=10.0$ selection pressure
    \item \textbf{Initial Solution:} Greedy baseline with constraint-aware selection
\end{itemize}

\textbf{Dataset Generation:}
\begin{itemize}
    \item \textbf{Base Samples:} 100 problems per seed (50 dense, 50 tree-structured)
    \item \textbf{Augmentation:} Each base sample expanded to 10 variations via:
    \begin{itemize}
        \item Problem instance variation (new problem, same type)
        \item Trace paraphrasing (same reasoning, different wording)
        \item Code reuse (same evolved code, new problem)
    \end{itemize}
    \item \textbf{Final Size:} 1000 samples per seed (100 evolved $\times$ 10 variations)
\end{itemize}

\textbf{System Prompt:}
We use a neutral SDS prompt that does not enumerate specific search strategies, while keeping the initializer, evolution workflow, model family, and search budget fixed. The evolution process uses:

\begin{tcolorbox}[colback=gray!5!white, colframe=gray!50!black, fonttitle=\bfseries, title=ShinkaEvolve System Prompt]
\scriptsize
You are an expert algorithmist specializing in combinatorial optimization, particularly the Synergistic Dependency Selection (SDS) problem.\\

The SDS problem requires selecting a subset of variables that:\\
1. Satisfies cardinality bounds (min/max number of items)\\
2. Respects mutex constraints (cannot select both items in a mutex pair)\\
3. Respects group constraints (at most one item per group)\\
4. Satisfies precedence constraints (if j is selected, i must be selected)\\
5. Maximizes the objective: sum of selected item weights + pairwise interaction values\\

The code should read JSON from stdin with "requirements" and "catalog", and output JSON to stdout with "selection" containing "variables" list.\\

Deceptive interactions can make purely greedy orderings unreliable, so prefer robust search procedures that still respect the execution-time budget.\\

Be creative and find efficient solutions that maximize the objective while respecting all constraints. The code must execute very quickly (within seconds) - prioritize time-efficient algorithms and avoid computationally expensive operations.
\end{tcolorbox}

\rev{\textbf{Comparison with Hero Prompt:} Unlike the Hero system prompt (Appendix~\ref{app:prompts}), which uses a scaffolded reasoning structure whose core loop is Deconstruct-Hypothesize-Critique to guide the Qwen-based policy, the ShinkaEvolve prompt is deliberately neutral and does not enumerate specific strategies such as greedy selection, dynamic programming, or backtracking. Since the ShinkaEvolve baseline uses strong proprietary reasoning models at test time (here GPT-5-mini and GPT-5-nano \citep{openai_gpt5_2025}), we use a constraint-focused prompt without Hero-specific reasoning steps and keep the initializer, workflow, and budget fixed. Under this fairer setup, the resulting ShinkaEvolve line is $91.6\%\pm1.2\%$ pass rate with $14.4\%\pm2.0\%$ mean feasible gap and $0.127\pm0.029$s mean per-instance solve cost, so the qualitative ordering from the main text remains unchanged even after neutralizing the prompt asymmetry.}

\textbf{Cost:} Each seed (100 base samples evolved) required approximately \$100 in OpenAI API calls for evolution and trace generation. The total cost for all three seeds (101, 202, 303) was approximately \$300.

\subsection{Baseline Implementation Details}
\label{app:baselines}

All baseline solvers are implemented in the \texttt{syndeopt} library and follow a unified API: \texttt{solver.solve(inst, budget\_sec, seed) $\rightarrow$ SolveResult}. We provide detailed specifications for each baseline:

\subsubsection{Greedy Solver}

The greedy solver (\texttt{GreedyMarginal}) uses a marginal-gain heuristic:

\begin{enumerate}
    \item \textbf{Initialization:} Start with empty selection $x = 0$
    \item \textbf{Upper Bound Phase:} While $|x| < U$:
    \begin{itemize}
        \item For each unselected variable $i$, compute marginal gain: $g_i = w_i + \sum_{j \in \text{selected}} W_{ij}$
        \item Select variable with maximum positive gain that maintains feasibility (excluding lower bound)
        \item Break if no positive gain available
    \end{itemize}
    \item \textbf{Lower Bound Phase:} While $|x| < L$:
    \begin{itemize}
        \item Select any feasible variable to satisfy cardinality lower bound
    \end{itemize}
    \item \textbf{Validation:} Return solution if feasible, otherwise return $-\infty$ score
\end{enumerate}

\textbf{Time Complexity:} $O(n^2)$ per iteration, bounded by $O(n \cdot U)$ total operations.

\subsubsection{Local Search Solver}

The local search solver (\texttt{LocalSearch1Flip}) implements a 1-flip hill-climber with random restarts:

\begin{enumerate}
    \item \textbf{Restart Strategy:} Default 20 restarts, scaled linearly with time budget: $\max(1, 20 \times \text{budget}/2.0)$
    \item \textbf{Initial Solution:} Random feasible solution (shuffled greedy construction)
    \item \textbf{Local Search:} For up to 2000 iterations per restart:
    \begin{itemize}
        \item Try all 1-flip neighbors (add/remove each variable)
        \item Accept best improving move (if gain $> 10^{-12}$)
        \item Stop if no improvement found
    \end{itemize}
    \item \textbf{Best Solution:} Track best solution across all restarts
\end{enumerate}

\textbf{Hyperparameters:} \texttt{restarts=20}, \texttt{iters=2000} (default values used in evaluation).

\subsubsection{Branch-and-Bound Solver}

The branch-and-bound solver (\texttt{BranchAndBound}) uses depth-first search with optimistic bounding:

\begin{enumerate}
    \item \textbf{Variable Ordering:} Sort variables by degree (descending) to prioritize high-connectivity variables
    \item \textbf{Bound Calculation:} For each node $(x, \text{idx})$:
    \begin{itemize}
        \item Current score: $S(x)$
        \item Optimistic bound: $S(x) + \sum_{\text{remaining}} \max(0, w_i + \sum_{j \in x} W_{ij})$
        \item Prune if bound $\leq$ best known score $+ 10^{-12}$
    \end{itemize}
    \item \textbf{Search Strategy:} Depth-first with stack-based recursion
    \item \textbf{Termination:} Time budget exceeded or search space exhausted
\end{enumerate}

\textbf{Pruning:} Uses optimistic bound based on remaining variables' maximum potential contribution.

\subsubsection{CP-SAT Solver}

The CP-SAT solver (\texttt{CPSATSolver}) uses Google OR-Tools' constraint programming solver \citep{perron_cp-sat-lp_2023}:

\begin{enumerate}
    \item \textbf{Model Construction:}
    \begin{itemize}
        \item Binary variables $x_i \in \{0,1\}$ for each variable
        \item Auxiliary variables $y_{ij} = x_i \land x_j$ for pairwise interactions
        \item Constraints: cardinality bounds, precedence, mutex, groups
    \end{itemize}
    \item \textbf{Objective:} Maximize $\sum_i w_i x_i + \sum_{(i,j)} W_{ij} y_{ij}$
    \item \textbf{Solver Configuration:}
    \begin{itemize}
        \item Time limit: \texttt{max\_time\_in\_seconds = budget\_sec}
        \item Random seed: \texttt{random\_seed = seed}
        \item Parallel workers: \texttt{num\_search\_workers = 8}
    \end{itemize}
    \item \textbf{Solution Tracking:} Callback records anytime solution trace
\end{enumerate}

\textbf{Note:} CP-SAT is an exact method in principle, but here it is used under the same time budget as the other baselines and should be read as the strongest time-limited classical reference. It uses multi-threaded search (8 workers) for parallel exploration.

\subsection{Base Model Scaling Analysis (Pass@k)}
\label{app:pass_at_k}

To rigorously quantify the latent capacity of the Base model (Section~\ref{sec:robustness}, Figure~\ref{fig:scaling_gap}), we employ a bootstrapped Pass@$k$ evaluation protocol. This methodology answers the counterfactual question: \textit{"If we were limited to only $k$ attempts per problem, what is the expected performance?"}

\paragraph{Data Generation.}
For the complete test set ($M=1000$ instances), we generate $N=64$ independent code solutions using the untrained Base model (Qwen2.5-Coder-14B-Instruct) with temperature $T=0.6$. Each solution is generated independently, ensuring no cross-contamination between samples. To ensure a fair comparison, the Base model is evaluated with the same system prompt as the Hero model (loaded from \texttt{config\_hero.yaml} and formatted using Qwen's chat template), providing identical instructions and scaffolding.

\paragraph{Bootstrap Estimation.}
For each scaling step $k \in \{1, 2, 4, 8, 16, 32, 64\}$, we perform a bootstrap analysis with $B=500$ iterations to estimate performance stability:
\begin{enumerate}
    \item \textbf{Sampling:} For each iteration $b \in [1, B]$ and problem $i$, we draw a random subsample $\mathcal{S}_{i,k}^b$ of size $k$ from the pool of $N=64$ generations \textit{without replacement}.
    \item \textbf{Metric Computation:} We compute the best-found metrics for the subsample:
    \begin{itemize}
        \item \textbf{Pass Rate:} $\mathbb{I}(\exists x \in \mathcal{S}_{i,k}^b : \text{is\_feasible}(x))$, where $\mathbb{I}$ is the indicator function.
        \item \textbf{Gap to VBS:} For feasible solutions, $\min_{x \in \mathcal{S}_{i,k}^b \cap \mathcal{F}} \text{Gap}(x)$, where $\mathcal{F}$ is the set of feasible solutions. If no feasible solution exists in the subsample, the gap is recorded as $1.0$ (100\%). The gap is computed as $\text{Gap}(x) = (\text{VBS} - \max(0, \text{score}(x))) / \text{VBS}$, where scores are clipped to non-negative values.
    \end{itemize}
    \item \textbf{Aggregation:} We compute the mean pass rate and mean gap to VBS across all $M=1000$ problems for each bootstrap iteration, then report the mean and standard deviation of these aggregate metrics across the $B=500$ bootstrap iterations.
\end{enumerate}

\paragraph{Interpretation of Variance.}
In Figure~\ref{fig:scaling_gap}, the error bars represent the \textit{standard error of the mean estimate} across bootstrap resamples, quantifying the statistical uncertainty of the aggregate performance metric. They do not represent the problem-to-problem variance (the spread of difficulty across the dataset), which is significantly larger and is visualized separately in the Robustness Profile (Figure~\ref{fig:robustness}). With $M=1000$ problems, the mean across problems is highly stable across bootstrap resamples, resulting in small error bars even when individual problems exhibit substantial performance variability.

\paragraph{Virtual Best Solver (VBS) Consistency.}
To ensure a fair comparison, the Virtual Best Solver (VBS) score used for calculating gaps is the global maximum across \textit{all} methods, including all baseline solvers (Greedy, Local Search, BnB, CP-SAT) and the union of all $N=64$ Base model samples. This prevents the Base model from artificially appearing optimal solely because it found a solution better than the deterministic baselines but worse than the strongest observed feasible score. The VBS is computed separately for each problem instance, considering only feasible solutions.

\paragraph{Best-of-N Collapsing.}
For the robustness profile comparison (Figure~\ref{fig:robustness}), we collapse the Base model's $N=64$ samples to a single ``Best-of-N'' solution per problem by selecting the best feasible solution (lowest gap to VBS). This enables a fair comparison with fine-tuned models, which generate a single solution per problem. The Pass@$k$ scaling analysis (Figure~\ref{fig:scaling_gap}) uses the raw $N=64$ samples and the unconditional SDS gap metric to demonstrate how performance scales with the number of attempts, while the main SDS table reports feasible-only mean gaps and the robustness profile uses the collapsed Best-of-N to show peak achievable performance.

\subsection{Universal Solver Search}
\label{app:universal_search}

\rev{To test whether a single code program could remain feasible across the entire 1,000-instance test split while approaching our Hero policy's 5.0\% gap, we conducted an adaptive tournament search across all unique codes generated by the Base model during Best-of-64 evaluation.}

\paragraph{Code Pool.}
From the Base model's Best-of-64 evaluation (Section~\ref{app:pass_at_k}), we extracted code blocks from all 64,000 generations per seed (3 seeds total). After deduplication via canonicalization (normalizing line endings and stripping trailing whitespace) and SHA-256 hashing, we obtained 191,699 unique codes across all seeds.

\paragraph{Tournament Methodology.}
We employed a two-stage adaptive tournament:
\begin{enumerate}
    \item \textbf{Stage 1 (Discriminative Filtering):} Evaluate all unique candidates on 30 missions selected from the test set (60\% hardest missions by Base model gap, 40\% random for coverage). Filter to candidates that are feasible and complete within timeout (5 seconds) on all 30 missions.
    \item \textbf{Stage 2 (Full Verification):} Evaluate top survivors (default: 10) on all 1,000 test missions.
\end{enumerate}

\paragraph{Evaluation Criteria.}
Candidates are evaluated using identical semantics as the main evaluation pipeline: same sandbox execution, constraint checking, and VBS reference (per-UUID from Base model's \texttt{metrics\_final.csv}). Optimality gap is computed as $(VBS - \max(0, \text{score})) / VBS$ for each mission independently, where scores are clipped to non-negative values.

\subsection{Raw Base-Code Audit}
\label{app:base_code_audit}

\rev{Because the main-text competence-wall argument now leans on properties of the raw Base code pool rather than on an ad hoc feasible-only subset, we audited exactly the same source used by universal search. The raw source consists of the three archived \texttt{generations.jsonl} files from Base Best-of-$64$ evaluation. Using the same extraction semantics as the universal-search pipeline---extract the \texttt{<code>} block, canonicalize whitespace, and deduplicate by content hash---we recover 192,000 raw generations, 191,699 extractable code blocks, and 191,699 unique codes.}

\paragraph{Audit Methodology.}
\rev{We first mark a code as \emph{SA-like} if it contains a temperature variable, a cooling schedule, and an exponential Metropolis-style acceptance expression. We then classify acceptance logic into four coarse cases: current-state acceptance, global-best acceptance, mixed signals, or unresolved by heuristic. Finally, for the non-global-best SA-like remainder, we apply an exclusive structural taxonomy based on four ingredients that matter for reusable SDS solvers: (i) explicit feasibility guards, (ii) explicit best-solution tracking, (iii) non-degenerate two-way neighborhood moves rather than one-sided edits, and (iv) non-ambiguous current-state acceptance. This is a structural audit, not a formal proof of semantic correctness, so we use it to characterize failure modes rather than to certify solver quality.}

\paragraph{Results.}
\rev{The Base model frequently retrieves SA-like syntax: 41,903 of the 191,699 unique codes (21.86\%) satisfy the SA-like heuristic. Within that SA-like subset, 12,050 codes (28.76\%) exhibit the global-best acceptance bug. Removing that bucket still leaves 29,853 SA-like codes, but most remain operationally incomplete: 42.00\% of the non-bug remainder lack explicit best-solution tracking, 14.52\% have ambiguous acceptance logic, and 14.34\% have weak or one-sided neighborhood structure. Only 23.79\% of the non-bug remainder appear structurally complete under this coarse audit. Table~\ref{tab:sa_failure_audit} summarizes the full exclusive split. This is why we interpret the global-best bug as one prominent failure mode rather than as a complete explanation: the broader pattern is incomplete operationalization of an SA template, which is consistent with the universal-search result that even the best reusable Base code remains far from Hero.}

\begin{table}[h]
\centering
\caption{\rev{Coarse structural audit of SA-like unique codes from the canonical raw Base pool (191,699 unique codes total; 41,903 SA-like). Percentages are computed within the SA-like subset, and the final column reports the corresponding percentage within the non-}\texttt{best\_bug}\rev{ remainder where relevant.}}
\label{tab:sa_failure_audit}
\small
\begin{tabular}{lccc}
\toprule
Bucket & Count & \% of SA-like & \% of non-bug remainder \\
\midrule
\texttt{best\_bug} & 12,050 & 28.76\% & -- \\
\texttt{ambiguous\_acceptance} & 4,336 & 10.35\% & 14.52\% \\
\texttt{current\_ok\_no\_guard} & 1,597 & 3.81\% & 5.35\% \\
\texttt{current\_ok\_no\_best\_tracking} & 12,538 & 29.92\% & 42.00\% \\
\texttt{current\_ok\_guarded\_but\_weak\_moves} & 4,281 & 10.22\% & 14.34\% \\
\texttt{current\_ok\_structurally\_complete} & 7,101 & 16.95\% & 23.79\% \\
\bottomrule
\end{tabular}
\end{table}

\subsection{Code and Data Availability}
\label{app:availability}

\rev{\textbf{Code Repository:} The full repository, including training, evaluation, reward, and dataset-generation code, is publicly available at \url{https://github.com/IDEALLab/neural-solver-synthesis}.}

The complete implementation, including:
\begin{itemize}
    \item Training scripts (GRPO configuration, SLURM job scripts)
    \item Evaluation pipeline (baseline solvers, metrics computation, plotting)
    \item Reward function implementations
    \item Dataset generation utilities
\end{itemize}

\rev{\textbf{Datasets:} The SDS and JSSP data bundles used in this work are provided through the Hugging Face artifact collection at \url{https://huggingface.co/collections/IDEALLab/neural-solver-synthesis}.}
\begin{itemize}
    \item \rev{Training datasets}
    \item \rev{Held-out test datasets}
    \item \rev{ShinkaEvolve comparison datasets}
\end{itemize}

\rev{\textbf{Model Checkpoints:} Hero and learned-ablation checkpoints are provided through the same Hugging Face collection at \url{https://huggingface.co/collections/IDEALLab/neural-solver-synthesis}.}
\begin{itemize}
    \item \rev{Hero and learned-ablation checkpoints}
\end{itemize}

\rev{\textbf{Experiment Artifacts:} Detailed logs, generated code, and training metrics are provided through the public code release and the accompanying Weights \& Biases project at \url{https://wandb.ai/neural-solver-synthesis/qwen-coder-sds-rl?nw=nwusersmassoudi-w}.}
\begin{itemize}
    \item \rev{Training runs}
    \item \rev{Evaluation results}
    \item \rev{Generated code samples}
\end{itemize}

\rev{\textbf{Container:} The GH200/ARM64 SDS training container used for the GRPO experiments is available on DockerHub at \url{https://hub.docker.com/r/soheylm/gh200-llm-sds-training/tags} with tag \texttt{25.01}.}

\rev{\textbf{ShinkaEvolve Solutions:} Pre-evolved code solutions for the comparison baseline are included in the Hugging Face artifact collection (see Section~\ref{app:shinkaevolve}): \url{https://huggingface.co/collections/IDEALLab/neural-solver-synthesis}.}

\section{Implementation Details}
\label{app:implementation}

\subsection{Code Structure}

\textbf{Codebase:} Our implementation is a fork of the \texttt{open-r1} repository \citep{openr1}. We utilize their distributed training harness and GRPO trainer integration, modifying the reward functions and system prompts as detailed in Section~\ref{sec:methodology}.

The implementation is organized as follows:
\begin{itemize}
    \item \texttt{rewards\_unified\_v2.py}: Main reward computation functions
    \item \texttt{evaluation/sds/generate.py}: Code generation using vLLM
    \item \texttt{evaluation/sds/evaluate.py}: Evaluation pipeline
    \item \texttt{evaluation/sds/utils.py}: Utility functions (code execution, constraint checking)
    \item \texttt{data/gen\_sds\_dataset.py}: Dataset generation
    \item \texttt{scripts/train\_*.slurm}: Training scripts
\end{itemize}

\subsection{Reproducibility Checklist}

To reproduce our results:
\begin{enumerate}
    \item Set environment variable \texttt{SEED} to one of \{101, 202, 303\}
    \item Generate dataset: \texttt{python data/gen\_sds\_dataset.py --seed \$SEED --num 10000}
    \item Run RL training (14B model, 3 nodes): \texttt{sbatch --nodes=3 --ntasks=3 scripts/train\_unified\_sds\_qwen\_coder.slurm --mode grpo\_cold --model 14B --seed \$SEED --config config\_hero.yaml}
    \item Evaluate standard SDS results: \texttt{sbatch scripts/eval\_sds\_pipeline.slurm --checkpoint-dir <PATH\_TO\_CHECKPOINT> grpo \$SEED}
    \item Evaluate fixed-code evidence when needed: \texttt{sbatch scripts/eval\_capstor\_sds\_fixed\_code.slurm <PATH\_TO\_SOLVER>}
    \item For ablation studies, replace \texttt{config\_hero.yaml} with \texttt{config\_ablation\_oracle.yaml}, \texttt{config\_ablation\_diversity.yaml}, \texttt{config\_ablation\_soft\_gate.yaml}, \texttt{config\_minimalist.yaml}, or \texttt{config\_ablation\_prompt.yaml}
\end{enumerate}

\paragraph{Note on Training Scripts.}
\rev{We use \texttt{train\_unified\_sds\_qwen\_coder.slurm} for the active SDS training path. The script supports multi-node training via \texttt{--nodes=3 --ntasks=3} for 14B models. For evaluation, \texttt{eval\_sds\_pipeline.slurm} covers the standard SDS report bundle and \texttt{eval\_capstor\_sds\_fixed\_code.slurm} covers fixed-code comparisons. Cluster-local wrappers and environment files can vary across systems, so the exact launcher layer may differ even when the underlying evaluator and configuration files remain the same.}

\paragraph{Configuration Files.}
All configuration files are located in \path{deps/open-r1/recipes/Qwen2.5-Coder-14B-Instruct/grpo/}:
\begin{itemize}
    \item \texttt{config\_hero.yaml}: parsimonious baseline (no oracle, no diversity)
    \item \texttt{config\_ablation\_oracle.yaml}: Adds oracle anchoring
    \item \texttt{config\_ablation\_diversity.yaml}: Adds diversity penalty
    \item \texttt{config\_ablation\_soft\_gate.yaml}: Replaces the hard nominal gate with a graded feasibility penalty
    \item \texttt{config\_ablation\_reward\_normalization.yaml}: Perturbs the SDS nominal reward normalization heuristic
    \item \texttt{config\_minimalist.yaml}: Removes structural scaffolding (uses minimal feasibility reward)
    \item \texttt{config\_ablation\_prompt.yaml}: Removes detailed system prompt (true ablation)
\end{itemize}
The config files specify reward functions, reward weights, system prompts, and all training hyperparameters. See Table~\ref{tab:ablation_configs} for a comparison of reward function differences.

\paragraph{Note on Bit-wise Reproducibility.}
While we fix the random seed for all experiments (data generation, initialization, and sampling), exact bit-wise reproducibility of training dynamics may not be achievable due to the non-deterministic nature of high-performance optimization kernels used on the GH200 hardware. Specifically:

\begin{itemize}
    \item \textbf{Flash Attention 2 Non-Determinism:} We utilize \texttt{flash\_attention\_2} for computational efficiency. To maximize parallelism, this implementation employs atomic operations (specifically atomic adds) during the backward pass. The execution order of these atomic operations is not guaranteed, leading to minor variations in gradient calculation ($\approx \pm 10^{-6}$) between runs.

    \item \textbf{ZeRO-3 Distributed Reduction:} Floating point addition is non-associative in BF16 precision (i.e., $(a+b)+c \neq a+(b+c)$). We utilize DeepSpeed ZeRO-3, which fully shards parameters and gradients across GPUs. The frequent \texttt{ReduceScatter} and \texttt{AllGather} operations required to reconstruct model states introduce accumulation noise dependent on network timing and reduction order across the cluster interconnect.

    \item \textbf{TensorFloat-32 (TF32):} We enable TF32 optimization to leverage the Hopper architecture's Tensor Cores. This truncates precision for matrix multiplications to accelerate throughput, introducing an additional layer of acceptable numerical noise compared to FP32.
\end{itemize}

We explicitly opted against enforcing strict determinism (e.g., via \texttt{torch.use\_deterministic\_algorithms(True)}) as it would preclude the use of these optimized kernels and result in prohibitive training costs (estimated $3\times$--$5\times$ slowdown).

\subsection{Dependency Versions}

\begin{table}[h]
\centering
\caption{Exact dependency versions for reproducibility.}
\label{tab:dependencies}
\small
\begin{tabular}{ll}
\toprule
Package & Version \\
\midrule
Base Container & NVIDIA PyTorch Release 25.01 \\
Python & 3.12 \\
CUDA & 12.8.0.038 \\
PyTorch & 2.6.0 \\
Transformers & 4.51.3 \\
vLLM & 0.8.0 \\
DeepSpeed & 0.16.7 \\
Syndeopt & Local (commit-based) \\
TRL & c04e84c4545acfaecdf7e0631ad07a86ab0fb2f6 \\
PEFT & 0.15.2 \\
NumPy & (from PyTorch container) \\
Pandas & (from PyTorch container) \\
\bottomrule
\end{tabular}
\end{table}

\subsection{Computational Complexity}

\paragraph{Reward Computation (Hero Configuration).} The time complexity of reward computation per sample for the Hero configuration:
\begin{itemize}
    \item \textbf{Code execution:} Linear in execution time, bounded by timeout $T=5$ seconds. Actual execution time depends on the generated code's algorithmic complexity (e.g., greedy heuristics are faster than simulated annealing with many iterations).
    \item \textbf{Score calculation:} $O(n + |W|)$ where $n$ is the number of variables and $|W|$ is the number of pairwise interactions. In practice, $|W| = O(n^2)$ for dense instances.
    \item \textbf{Constraint checking:} $O(n + m)$ where $n$ is the number of variables and $m$ is the total number of constraints (precedence pairs, mutex pairs, group constraints). For typical instances, $m = O(n)$.
    \item \textbf{Format/structure validation:} $O(L)$ where $L$ is the length of the generated code (typically $< 1000$ tokens).
    \item \textbf{Total per sample:} Dominated by code execution time, with overhead of $O(n + m + |W|)$ for constraint checking and score calculation.
\end{itemize}

\paragraph{Ablation-Specific Complexity.} Additional computational overhead for ablation configurations:
\begin{itemize}
    \item \textbf{+ Oracle Anchoring:} Adds $O(n \log n)$ greedy solver computation per sample (used for normalization baseline).
    \item \textbf{+ Diversity Penalty:} Adds $O(G \cdot L)$ computation per batch to compute 4-gram Jaccard similarity across $G=64$ samples of average length $L$.
\end{itemize}

\textbf{Note on Timeout:} Both training and evaluation use a 5-second execution time limit per code sample. Solutions that exceed this limit are marked as timeouts. During training, timeouts receive zero reward (infeasible), encouraging the model to generate efficient code. During evaluation, timeouts are tracked separately from constraint violations (see Table~\ref{tab:error_types}).

\paragraph{Training Time.} 
\begin{itemize}
    \item Global batch size: 4 prompts per step (8 training GPUs $\times$ 8 generations/GPU $\times$ 4 gradient accumulation = 256 generation slots, divided by 64 generations per prompt)
    \item Completions per step: 256 (4 prompts $\times$ 64 generations)
    \item Training samples seen: 360 prompts at checkpoint 90 (0.045 epoch), or $\approx$ 8,000 prompts for full epoch ($\approx$ 2,000 steps)
    \item Total training time: 4 hours per experiment (standard ablations)
    \item Evaluation: $\approx$ 40 minutes--1 hour (for 1000 test problems: LLM inference, code execution, and baseline solvers using 32 parallel workers)
\end{itemize}

\subsection{General Coding Evaluation}
For the catastrophic forgetting analysis (Table~\ref{tab:bigcode_results}), we utilized the \textbf{BigCode Evaluation Harness}. To ensure a fair comparison with the Qwen2.5-Coder technical report \citep{hui_qwen25-coder_2024}, we employed the following settings:
\begin{itemize}
    \item \textbf{Decoding:} Greedy (Temperature $= 0$, Top-p $= 1.0$)
    \item \textbf{Samples:} $N=1$ (Pass@1)
    \item \textbf{Tasks:} 
    \begin{itemize}
        \item \textbf{HumanEval:} 164 hand-written Python programming problems \citep{chen_evaluating_2021}.
        \item \textbf{MBPP (Sanitized):} 500 crowd-sourced Python programming problems \citep{austin_program_2021}.
\end{itemize}
    \item \textbf{Qwen-Specific Configuration:} We applied the workaround from BigCode issue \#308 to preserve trailing newlines in prompts (required for proper indentation in Qwen Coder models) and removed \verb|\n#| and \verb|\nif| from stop words to prevent early stopping when models add comments or \verb|if __name__ == __main__:| blocks.
    \item \textbf{Prompt Formatting:} The BigCode harness uses raw prompts without chat template formatting. This may differ from Qwen's internal evaluation methodology (which likely applies their chat template), potentially explaining slight discrepancies between our base model results and those reported in the Qwen technical report.
\end{itemize}
Values reported for fine-tuned models are the mean and standard deviation across the 3 independent training seeds (101, 202, 303).

\section{Additional Results}
\label{app:results}

\subsection{Error Analysis Discussion.}

\begin{figure}[h]
\centering
\includegraphics[width=0.7\textwidth]{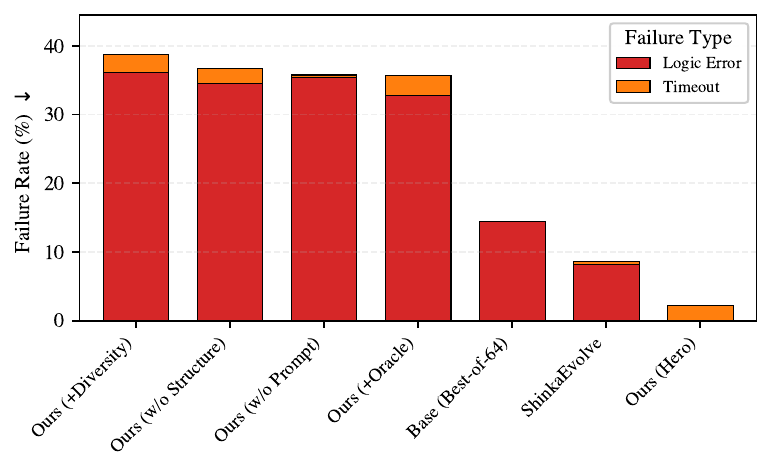}
\caption{\textbf{Failure Mode Analysis.} Stacked bar chart showing Logic Errors (red) vs. Timeouts (orange) for our methods and ShinkaEvolve. Hero achieves near-zero logic errors, while ablations collapse primarily due to constraint violations, not timeouts.}
\label{fig:failure_modes}
\end{figure}

\begin{table}[h]
\centering
\caption{\rev{Diagnostic failure-mode breakdown} across all methods (Mean $\pm$ Std across 3 seeds). The Hero configuration achieves 97.8\% success with only 2.2\% timeouts and zero logic errors. Ablations show dramatically higher constraint violation rates, while the ``w/o Prompt'' ablation uniquely suffers from 29.8\% runtime errors. \rev{For learned methods, the error columns are diagnostic tags rather than mutually exclusive outcome categories, so rows can sum above 100\%. For classical baselines, the ``Constraint Violation'' column is a coarse non-pass bucket in the current evaluator because detailed error types are not recorded for those methods.}}
\label{tab:error_types}
\small
\resizebox{\columnwidth}{!}{
\begin{tabular}{lcccccc}
\toprule
Method & None (Valid) (\%) $\uparrow$ & Timeout (\%) $\downarrow$ & Constraint Violation (\%) $\downarrow$ & Syntax Error (\%) $\downarrow$ & Runtime Error (\%) $\downarrow$ & JSON Parse Error (\%) $\downarrow$ \\
\midrule
Ours (Hero) & $97.8_{\pm 0.2}$ & $2.2_{\pm 0.2}$ & $\mathbf{0.0}$ & $\mathbf{0.0}$ & $\mathbf{0.0}$ & $\mathbf{0.0}$ \\
Base (Best-of-64) & $85.6_{\pm 1.3}$ & $\mathbf{0.0}$ & $14.4_{\pm 1.3}$ & $\mathbf{0.0}$ & $\mathbf{0.0}$ & $\mathbf{0.0}$ \\
Ours (+Oracle) & $65.3_{\pm 11.5}$ & $2.9_{\pm 0.6}$ & $31.7_{\pm 11.8}$ & $\mathbf{0.0}$ & $3.2$ & $\mathbf{0.0}$ \\
Ours (+Diversity) & $61.3_{\pm 9.8}$ & $2.6_{\pm 0.5}$ & $36.1_{\pm 9.5}$ & $\mathbf{0.0}$ & $\mathbf{0.0}$ & $\mathbf{0.0}$ \\
Ours (w/o Structure) & $74.8_{\pm 19.2}$ & $2.2_{\pm 0.9}$ & $34.5_{\pm 4.0}$ & $\mathbf{0.0}$ & $\mathbf{0.0}$ & $\mathbf{0.0}$ \\
Ours (w/o Prompt) & $64.4_{\pm 7.8}$ & $0.4_{\pm 0.4}$ & $25.4_{\pm 19.4}$ & $\mathbf{0.0}$ & $29.8$ & $\mathbf{0.0}$ \\
ShinkaEvolve & $91.6_{\pm 1.9}$ & $0.4$ & $5.7_{\pm 0.5}$ & $0.2$ & $0.5_{\pm 0.5}$ & $3.0_{\pm 1.4}$ \\
CP-SAT & $\mathbf{98.3_{\pm 0.3}}$ & $\mathbf{0.0}$ & $1.7_{\pm 0.3}$ & $\mathbf{0.0}$ & $\mathbf{0.0}$ & $\mathbf{0.0}$ \\
Local Search & $\mathbf{98.3_{\pm 0.3}}$ & $\mathbf{0.0}$ & $1.7_{\pm 0.3}$ & $\mathbf{0.0}$ & $\mathbf{0.0}$ & $\mathbf{0.0}$ \\
Greedy & $\mathbf{98.3_{\pm 0.3}}$ & $\mathbf{0.0}$ & $1.7_{\pm 0.3}$ & $\mathbf{0.0}$ & $\mathbf{0.0}$ & $\mathbf{0.0}$ \\
BnB & $97.3_{\pm 0.4}$ & $\mathbf{0.0}$ & $2.7_{\pm 0.4}$ & $\mathbf{0.0}$ & $\mathbf{0.0}$ & $\mathbf{0.0}$ \\
\bottomrule
\end{tabular}
}
\end{table}

\rev{As a diagnostic complement to the configuration ablations and normalization sensitivity study,} Table~\ref{tab:error_types} provides a comprehensive breakdown of error types across all methods, aggregated across three seeds. The Hero configuration achieves a 97.8\% success rate with only 2.2\% timeouts and zero logic errors (syntax, runtime, or constraint violations), demonstrating that the feasibility-gated reward structure successfully guides the model to generate both valid and efficient code.

\textbf{Hero Configuration:} The near-perfect performance (97.8\% valid, 2.2\% timeouts, 0.0\% logic errors) validates our parsimonious approach. \rev{Across the three seeds, 57 of the 65 timeout cases occur on Hard instances and the remaining 8 occur on Moderate instances, with none on Trivial instances. More quantitatively, the timeout rows are significantly larger and denser than the non-timeout rows: they have mean $n_{\text{vars}}=74.3$ versus $22.8$ and mean interaction count $1518.7$ versus $153.8$, with Mann-Whitney $p<10^{-40}$ for both comparisons; all 65 timeout cases fall in the top decile of both instance size and interaction count.} The small timeout rate therefore concentrates on the hardest SDS strata rather than on easy cases. \rev{The saved timeout rows do not contain completed selections or violation counts, so the current evidence supports the interpretation that these failures reflect search cost on large, interaction-dense hard instances, but it does not yet isolate a single inner-loop mechanism such as retry-loop behavior.}

\textbf{Ablation Failure Modes:} The error breakdown reveals distinct failure patterns for each ablation. Oracle Anchoring and Diversity ablations exhibit 31.7\% and 36.1\% constraint violation rates respectively, indicating that these reward modifications cause the model to generate invalid solutions rather than slow but correct ones. Similarly, the ``w/o Structure'' ablation yields a 34.5\% violation rate. This is consistent with the view that without structural scaffolding, the model more often falls back to \textit{brittle} passive filtering logic rather than synthesizing the necessary \textit{robust constraint verification}. Finally, the ``w/o Prompt'' ablation uniquely suffers from 29.8\% runtime errors. We interpret this as an ablation of the full deployed prompt package rather than a perfectly isolated cognitive-scaffolding intervention, since the minimal prompt also removes some environment reminders; the result is therefore consistent with algorithmic scaffolding in the system prompt being important for producing executable Python code.

\textbf{Baseline Comparison:} \rev{For classical baselines, the reported ``Constraint Violation'' rate should be read as a coarse non-pass rate under the evaluation harness, not as evidence that a solver such as CP-SAT explicitly returned an invalid assignment. Baseline rows only preserve success/failure, so any non-pass baseline case is mapped into that column in the current table. In the SDS runs used here, the CP-SAT, Local Search, and Greedy non-pass cases occur only on Hard instances and are recorded with no valid returned solution (score $=-\infty$).} BnB achieves 97.3\% success with 2.7\% non-pass cases and additionally differs in that its failures align with the 5-second budget. ShinkaEvolve achieves 91.6\% success with 5.7\% constraint violations and 3.0\% JSON parse errors, indicating that test-time evolution can occasionally produce malformed outputs.

The stark contrast between Hero (0.0\% logic errors) and ablations (25.4\%--36.1\% constraint violations) demonstrates that adding complexity fundamentally breaks constraint satisfaction, validating our parsimonious hypothesis.

\subsection{General Coding Robustness}

\begin{table}[h]
\centering
\caption{\textbf{General Coding Robustness (Pass@1).} Evaluation on HumanEval and MBPP (Greedy Decoding) across 3 seeds. Our methods show no statistically significant degradation compared to the Base model, indicating that the RL process avoids catastrophic forgetting.}
\label{tab:bigcode_results}
\begin{tabular}{lcc}
\toprule
Method & HumanEval (\%) $\uparrow$ & MBPP (\%) $\uparrow$ \\
\midrule
Base & $\mathbf{82.3}$ & $74.8$ \\
Ours (Hero) & $80.9_{\pm 0.9}$ & $74.4_{\pm 0.2}$ \\
Ours (+Oracle) & $80.5_{\pm 1.1}$ & $74.3_{\pm 0.9}$ \\
Ours (+Diversity) & $80.3_{\pm 1.4}$ & $\mathbf{74.9_{\pm 0.1}}$ \\
Ours (w/o Structure) & $80.9_{\pm 0.9}$ & $73.7_{\pm 0.1}$ \\
Ours (w/o Prompt) & $81.5_{\pm 0.9}$ & $73.8_{\pm 0.8}$ \\
\bottomrule
\end{tabular}
\end{table}

\subsection{Best Single-Code Search Results}

\rev{Table~\ref{tab:universal_search} summarizes the best single-code search results across 191,699 unique candidates from 3 seeds. This appendix analysis complements the competence-wall claim by asking whether exhaustive base-model sampling can recover one reusable solver that approaches Hero.}

While the best single code achieved 98.4\% feasible rate (comparable to Hero's 97.8\%), the best candidate (from seed 101) achieved a 24.04\% mean gap to VBS, which is nearly 5× worse than our Hero policy's 5.0\% gap to VBS. This demonstrates that even the best single code from 64k samples cannot match the reliability and performance of our RL-trained Hero policy, which synthesizes a specialized solver through training rather than sampling.

\begin{table}[h]
\centering
\caption{Best Single-Code Search Results. Best candidate performance compared to Hero policy.}
\label{tab:universal_search}
\small
\begin{tabular}{lcc}
\toprule
Metric & Best Single Code & Hero Policy \\
\midrule
Mean Optimality Gap & 24.04\% & 5.0\% \\
Feasible Rate & 98.4\% & 97.8\% \\
Timeout Rate & 1.6\% & 2.2\% \\
\bottomrule
\end{tabular}
\end{table}

\section{Qualitative Analysis}
\label{app:qualitative}

\subsection{Algorithmic Convergence and Reasoning Traces}
\rev{To investigate the generated behaviors of the policy, we analyzed the internal monologue and emitted code. As shown in Figure~\ref{fig:emergent_code}, the model (Seed 303) explicitly identifies the risk of local optima in the \texttt{<think>} block and operationalizes a Simulated Annealing strategy to mitigate it. This qualitative example supports the main-text semantic-repair story by showing how the scaffold is instantiated in executable form.}

\begin{figure}[t!]
\centering
\begin{tcolorbox}[colback=blue!5!white, colframe=blue!50!black, title=\textbf{Reasoning Trace (Seed 303)}, fonttitle=\bfseries, arc=0mm]
\scriptsize
\textbf{Problem Deconstruction:} The SDS problem involves selecting a subset of variables to maximize total value, respecting constraints (cardinality, precedence, mutex, groups).\\
\textbf{Hypothesis:} Given the complexity and the need to respect constraints, a simulated annealing approach seems suitable due to its ability to explore the solution space and escape local optima.\\
\textbf{Implementation Plan:} Accept the new solution if it improves the objective function, or accept with a certain probability if it decreases (to escape local optima). Cool the temperature according to a cooling schedule.
\end{tcolorbox}
\begin{tcolorbox}[colback=gray!5!white, colframe=gray!50!black, title=\textbf{Generated Solution (Seed 303)}, fonttitle=\bfseries, arc=0mm, top=0mm]
\scriptsize
\begin{verbatim}
# ... [Initialization of T=1000, cooling=0.99] ...
for _ in range(n_iterations):
    # Neighbor generation (1-flip). Renamed
    # 'neighbor_solution' to 'neighbor' for brevity
    neighbor = current_sol[:]
    idx = random.randint(0, n_vars - 1)
    neighbor[idx] = not neighbor[idx]
    
    # Constraint Guard (Rejection Sampling)
    while not is_feasible(neighbor):
        idx = random.randint(0, n_vars - 1)
        neighbor[idx] = not neighbor[idx]
    
    # Metropolis Acceptance Logic
    n_score = calculate_score(neighbor)
    delta = n_score - current_score
    
    # Use backslash for clean line continuation
    if delta > 0 or \
       random.random() < math.exp(delta/T):
            current_sol = neighbor 
        current_score = n_score
    
    T *= cooling_rate
\end{verbatim}
\end{tcolorbox}
\caption{\textbf{Generated Meta-Heuristic.} The model (Seed 303) identifies the risk of local optima and implements a robust Simulated Annealing solver to mitigate it. The ``Constraint Guard'' loop (lines 11-14) is a critical generated feature that ensures search efficiency.}
\label{fig:emergent_code}
\end{figure}

\rev{To ensure our findings are not artifacts of a single training run, we analyzed the generated code across all three independent seeds (101, 202, 303). While all seeds converged to the same \textbf{Simulated Annealing} strategy class (99.8\% convergence rate), they instantiated it with distinct implementation flavors.}

\paragraph{Convergence Analysis Methodology.}
\rev{We computed the convergence rate using static code analysis on all feasible solutions ($N=2,935$ across three seeds). For each solution, we extracted hyperparameters (temperature $T$, cooling rate $\alpha$, iteration count) via regex patterns and detected structural components: a \textbf{constraint guard} (rejection sampling pattern \texttt{while not is\_feasible()}) and the \textbf{Metropolis criterion} (probabilistic acceptance \texttt{exp($\Delta$/T)}). A solution is classified as matching the Simulated Annealing template if it contains both structural components and valid hyperparameters ($T \geq 100$, $0.8 \leq \alpha < 1.0$, iterations $\geq 100$ or dynamic termination). This structural matching approach (rather than exact hyperparameter matching) reflects that different seeds converged to different but valid local optima while instantiating the same algorithm class.}

\subsection{Cross-Seed Implementation Diversity}
\rev{While all seeds converged to the same \textbf{Simulated Annealing} strategy class (99.8\% convergence rate), they instantiated it with distinct implementation flavors. This variance supports the view that the model is learning an algorithmic \textit{concept} rather than memorizing a single text snippet.}

\paragraph{Hyperparameter Diversity.}
\rev{As shown in Table~\ref{tab:seed_comparison}, the seeds converged to different local optima for the cooling schedule. Seed 303 favored a faster schedule ($\alpha=0.99$) with fixed iterations ($n=1000$), while Seed 101 and 202 instantiated a ``deeper search'' strategy ($\alpha=0.995$) with higher iteration counts (Seed 101: $n=10000$) or dynamic loop conditions (Seed 202: \texttt{while T > 1}). This variance supports the view that the model is learning a robust algorithmic \textit{concept} that can be instantiated in multiple valid ways.}

\begin{table}[h]
\centering
\caption{\textbf{Implementation Variations Across Seeds.} All seeds converged to Simulated Annealing, but tuned the parameters differently.}
\label{tab:seed_comparison}
\small
\begin{tabular}{lccc}
\toprule
Feature & Seed 101 & Seed 202 & Seed 303 \\
\midrule
Algorithm & Sim. Annealing & Sim. Annealing & Sim. Annealing \\
Temperature ($T_0$) & $1000$ & $1000$ & $1000$ \\
Cooling Rate ($\alpha$) & $0.995$ & $0.995$ & $0.99$ \\
Iteration Logic & Fixed (10k) & Dynamic ($T > 1$) & Fixed (1k) \\
Constraint Guard & \checkmark & \checkmark & \checkmark \\
\bottomrule
\end{tabular}
\end{table}

\paragraph{Full Implementation (Seed 202).}
Below is the complete, raw code generated by Seed 202. Unlike the fixed \texttt{for} loop in Seed 303 (Main Text), this version utilizes a dynamic \texttt{while T > 1} termination condition and encapsulates the logic in a clean functional structure. This demonstrates the model's capability to generate complete, executable solvers with varying control flow structures.

\begin{tcolorbox}[colback=gray!5!white, colframe=gray!50!black, fonttitle=\bfseries, arc=0mm, title=\textbf{Full Solution (Seed 202)}, breakable]
\scriptsize
\begin{verbatim}
import sys
import random
import math
import json

def solve_sds():
    # Read input
    input_data = sys.stdin.read()
    data = json.loads(input_data)
    
    requirements = data["requirements"]
    catalog = data["catalog"]
    
    n_variables = requirements["n_variables"]
    min_card, max_card = requirements["cardinality_bounds"]
    precedence = requirements["precedence"]
    mutex = requirements["mutex"]
    groups = requirements["groups"]
    variables = catalog["variables"]
    interactions = catalog["interactions"]
    
    # Function to calculate the total value of a selection
    def calculate_value(selection):
        total_value = 0
        for idx in selection:
            total_value += variables[idx]["weight"]
        for i in range(len(selection)):
            for j in range(i + 1, len(selection)):
                interaction_key = f"{min(selection[i], selection[j])},{max(selection[i], selection[j])}"
                if interaction_key in interactions:
                    total_value += interactions[interaction_key]
        return total_value
    
    # Function to check feasibility
    def is_feasible(selection):
        # Precedence
        for i, j in precedence:
            if j in selection and i not in selection:
                return False
        # Mutex
        for a, b in mutex:
            if a in selection and b in selection:
                return False
        # Groups
        for group_id, group_vars in groups.items():
            selected_in_group = [var for var in group_vars if var in selection]
            if len(selected_in_group) > 1:
                return False
        # Cardinality
        if not (min_card <= len(selection) <= max_card):
            return False
        return True
    
    # Simulated Annealing
    initial_selection = random.sample(list(range(n_variables)), random.randint(min_card, max_card))
    while not is_feasible(initial_selection):
        initial_selection = random.sample(list(range(n_variables)), random.randint(min_card, max_card))
    
    current_selection = initial_selection
    current_value = calculate_value(current_selection)
    T = 1000
    cooling_rate = 0.995
    
    # Dynamic termination based on Temperature
    while T > 1:
        # Propose a new selection
        new_selection = current_selection[:]
        if random.random() < 0.5 and len(new_selection) < max_card:
            candidate = random.choice(list(set(range(n_variables)) - set(new_selection)))
            new_selection.append(candidate)
        elif len(new_selection) > min_card:
            candidate = random.choice(new_selection)
            new_selection.remove(candidate)
        
        # Constraint Guard
        if is_feasible(new_selection):
            new_value = calculate_value(new_selection)
            delta = new_value - current_value
            # Metropolis Criterion
            if delta > 0 or random.random() < math.exp(delta / T):
                current_selection = new_selection
                current_value = new_value
        
        T *= cooling_rate
    
    # Output the best found selection
    print(json.dumps({"selection": {"variables": current_selection}}))

if __name__ == "__main__":
    solve_sds()
\end{verbatim}
\end{tcolorbox}

\subsection{Comparative Analysis: The Implementation Wall}
\label{app:implementation_wall}

To rigorously validate the ``Implementation Wall'' hypothesis, we provide a direct code comparison between the Minimalist (w/o Structure ablated) policy and the Hero policy. Both policies attempted to implement Simulated Annealing, but they differ fundamentally in how they handle invalid states during the search loop.

\paragraph{The Passive Filtering Failure (Minimalist Policy).}
The code below (Seed 303, Minimalist Ablation) illustrates passive constraint filtering.
\textbf{The Flaw:} The cooling schedule executes regardless of whether a valid neighbor was found. If \texttt{is\_valid\_solution} returns \texttt{False}, the iteration is wasted, and the temperature decreases. In dense constraint landscapes, this leads to a ``frozen'' search where the temperature drops to zero before the agent has successfully explored the landscape.

\begin{tcolorbox}[colback=red!5!white, colframe=red!50!black, title=\textbf{Minimalist Failure (Passive Filtering)}, fonttitle=\bfseries]
\scriptsize
\begin{verbatim}
# ... [Setup code omitted] ...

    # Simulated annealing loop
    while temperature > 1e-10:
        # 1. Propose a new solution (Perturbation)
        new_solution = current_solution[:]
        # ... [Random mutation logic] ...

        # 2. PASSIVE CHECK: Only proceed if valid
        if is_valid_solution(new_solution):
            new_score = calculate_score(new_solution)
            
            # Metropolis Acceptance
            if new_score > current_score or \
               random.uniform(0, 1) < math.exp((new_score - current_score) / temperature):
                current_solution = new_solution
                current_score = new_score
                if current_score > best_score:
                    best_solution = current_solution
                    best_score = current_score

        # 3. CRITICAL FLAW: Cool down regardless of validity
        # If the mutation above was invalid, we just wasted 
        # a step of our cooling schedule.
        temperature *= cooling_rate

# ... [Output logic] ...
\end{verbatim}
\end{tcolorbox}

\paragraph{The Active Guard Success (Hero Policy).}
In contrast, the Hero policy (Seed 303) implements an active constraint guard.
\textbf{The Fix:} The policy uses a \texttt{while} loop to enforce validity \textit{before} proceeding to the Metropolis step. This ensures that every decrement of the temperature corresponds to a valid evaluation of the objective landscape.

\begin{tcolorbox}[colback=green!5!white, colframe=green!50!black, title=\textbf{Hero Success (Active Guard)}, fonttitle=\bfseries]
\scriptsize
\begin{verbatim}
# ... [Setup code omitted] ...

    for _ in range(n_iterations):
        # 1. Propose a new solution
        neighbor = current_sol[:]
        idx = random.randint(0, n_vars - 1)
        neighbor[idx] = not neighbor[idx]
        
        # 2. ACTIVE GUARD: Retry until valid
        # The search does not proceed until a feasible 
        # state is found.
        while not is_feasible(neighbor):
            idx = random.randint(0, n_vars - 1)
            neighbor[idx] = not neighbor[idx]
        
        # 3. Metropolis Acceptance
        n_score = calculate_score(neighbor)
        delta = n_score - current_score
        
        if delta > 0 or random.random() < math.exp(delta/T):
            current_sol = neighbor 
            current_score = n_score
        
        # 4. Cool down only after a valid step attempts
        T *= cooling_rate
\end{verbatim}
\end{tcolorbox}

\subsection{Base Model Failure: Logic Hallucination}
\label{app:base_failure}

While the Base Model frequently retrieves Simulated Annealing-like syntax (21.9\% of unique raw codes in the canonical audit), it fails to operationalize the physics correctly. The code below (extracted from a high-scoring Base Model sample) demonstrates the \textbf{Global Best Bug}.

\textbf{The Flaw:} The acceptance probability $P = \exp((S_{\text{new}} - S_{\text{best}})/T)$ compares the neighbor to the running global best found so far, rather than the \textit{current} state. This causes the acceptance probability to vanish rapidly once the search reaches a decent local maximum under the current neighborhood, effectively freezing the search and preventing the algorithm from accepting downhill moves needed to escape traps.

\begin{tcolorbox}[colback=orange!5!white, colframe=orange!50!black, title=\textbf{Base Model Failure (Logic Hallucination)}, fonttitle=\bfseries]
\scriptsize
\begin{verbatim}
# ... [Standard setup matches Hero] ...

    # Simulated Annealing Loop
    for _ in range(max_iterations):
        # ... [Neighbor generation with Active Guard] ...
        
        # Calculate value
        neighbor_value = calculate_value(neighbor_solution, variables, interactions)
        
        # LOGIC HALLUCINATION:
        # Compares neighbor against BEST_value, not CURRENT_value.
        # This breaks the annealing physics.
        if neighbor_value > best_value or \
           random.random() < math.exp((neighbor_value - best_value) / temperature):
            
            solution = neighbor_solution[:] # Moves current pointer
            if neighbor_value > best_value:
                best_value = neighbor_value # Updates best
                
        # Cool down
        temperature *= cooling_rate
\end{verbatim}
\end{tcolorbox}


\end{document}